\documentclass[runningheads]{llncs}

\usepackage{eccv}

\usepackage{eccvabbrv}

\usepackage{graphicx}
\usepackage{booktabs}
\usepackage[table]{xcolor}
\definecolor{polyRed}{HTML}{E53935}
\definecolor{lineYellow}{HTML}{FBC02D}
\definecolor{boxGreen}{HTML}{81C784}
\definecolor{pointBlue}{HTML}{64B5F6}
\usepackage{multirow}
\usepackage{pifont}
\usepackage{mathtools}
\usepackage{makecell}
\newcommand{\cmark}{\ding{51}}
\newcommand{\xmark}{\ding{55}}
\usepackage[accsupp]{axessibility}

\usepackage[pagebackref,breaklinks,colorlinks,citecolor=eccvblue]{hyperref}

\usepackage{orcidlink}
\usepackage{textcomp,xspace}
\newcommand{\TerraDiTOmega}{TerraDiT-\texorpdfstring{$\mathrm{\Omega}$}{Omega}\xspace}
\newcommand{\TextOmega}{\texorpdfstring{$\mathrm{\Omega}$}{Omega}}

\usepackage{tabto}
\usepackage{algorithm}
\usepackage{algorithmic}
\newcommand{\beginsupplement}{
    \clearpage
    \setcounter{table}{0}
    \renewcommand{\thetable}{S\arabic{table}}
    \setcounter{figure}{0}
    \renewcommand{\thefigure}{S\arabic{figure}}
    \setcounter{section}{0}
    
    \setcounter{footnote}{0} 
    
    \title{\TerraDiTOmega: Unified Spatial Control for Satellite Image Synthesis with Any Geospatial Primitive \\[0.3em] \normalsize -- Supplementary Material --}
    \titlerunning{\TerraDiTOmega{} -- Supplementary Material}

    \author{}
    \authorrunning{}
    \institute{}
    \email{}
    \maketitle
}

\makeatletter
\renewcommand\subsubsection{\@startsection{subsubsection}{3}{\z@}
                       {-6pt \@plus -2pt \@minus -1pt}
                       {-0.5em}
                       {\normalfont\normalsize\bfseries}}
\makeatother

\makeatletter
\newcommand{\printfnsymbol}[1]{%
  \textsuperscript{\@fnsymbol{1}}%
}
\makeatother

\begin{document}

\title{\TerraDiTOmega: Unified Spatial Control for Satellite Image Synthesis with Any Geospatial Primitive}

\titlerunning{\TerraDiTOmega}

\author{Brian Wei\thanks{\scriptsize Equal contribution.} \and
Srikumar Sastry\printfnsymbol{1}  \and
Daniel Cher\printfnsymbol{1}  \and
Eric Xing\and
Nathan Jacobs }

\authorrunning{B.~Wei et al.}

\institute{Washington University in St. Louis \\
\email{\{b.j.wei, s.sastry, cher, e.xing, jacobsn\}@wustl.edu}}

\maketitle

\begin{figure}[b!]
    \centering
    \includegraphics[width=0.7105\linewidth]{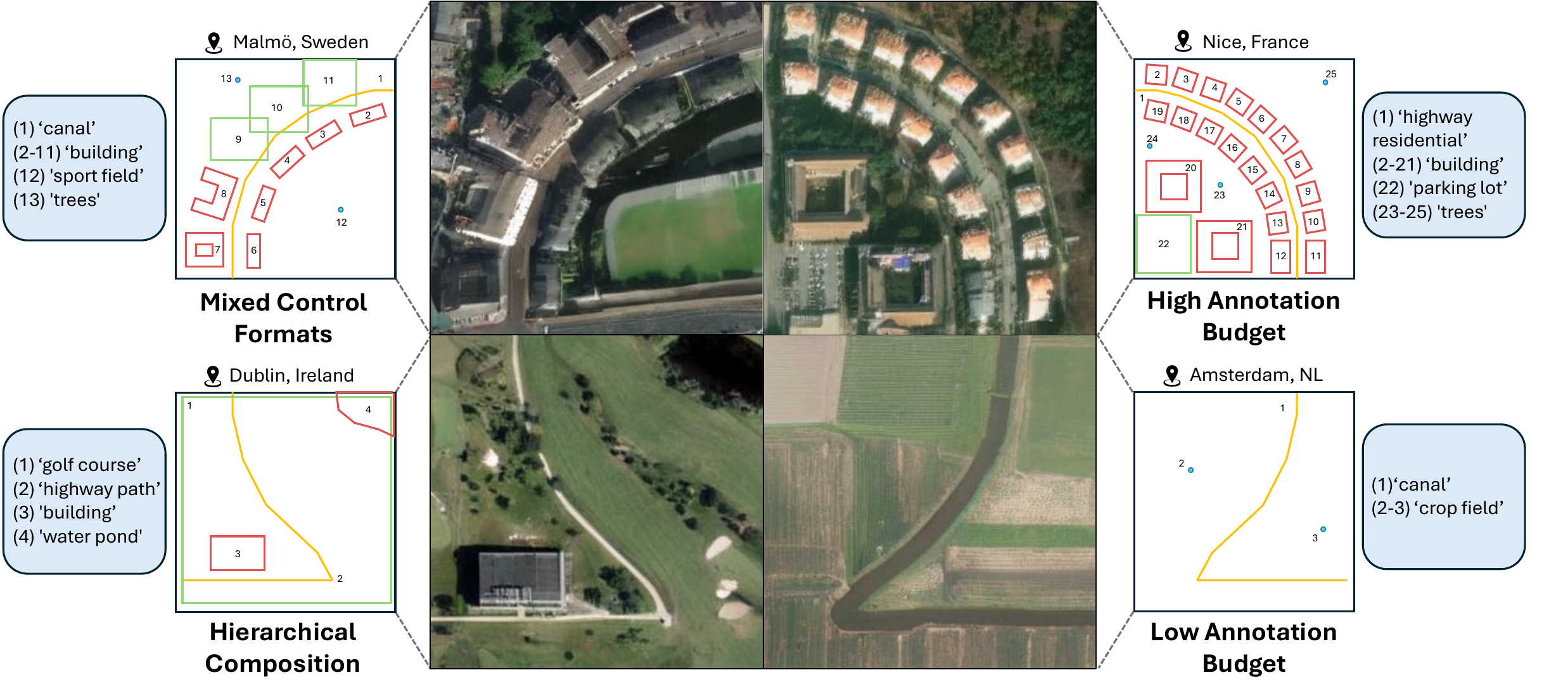}
    \caption{\TerraDiTOmega{} enables precise synthesis using \textcolor{polyRed}{polygons}, \textcolor{lineYellow}{polylines}, \textcolor{boxGreen}{bounding boxes}, \textcolor{pointBlue}{points}, along with text and geolocation conditioning.}
    \label{fig:teaser}
\end{figure}

\begin{abstract}

Generative models have achieved remarkable progress, yet applying them to satellite imagery remains challenging. Unlike natural imagery, satellite scenes are structured by spatially complex and semantically distinct geometries. Prior work addresses this complexity by adapting natural image frameworks using dense rasters or sparse prompts, trading off annotation cost and fidelity while breaking compatibility with vector primitives commonly used to represent geographic information. We introduce \TerraDiTOmega, a unified spatial control framework that generates satellite imagery directly from any native geospatial primitive. By jointly leveraging precise annotations (polygons, polylines) and coarser ones (bounding boxes, points), the model supports controllable layouts across varying annotation budgets, broadening applicability to design tasks such as urban planning while remaining naturally compatible with end-to-end GeoAI workflows. To effectively leverage these primitives during generation, we propose Geometry-Aware Local Attention, a conditioning mechanism that injects explicit geometric cues into the attention space. Across all conditioning formats, our approach consistently outperforms both dense-control and sparse-control baselines. Furthermore, this flexibility enables controllable synthetic data augmentation using a single generative model, improving downstream performance on land-cover segmentation, object detection, road graph extraction, and scene classification. Code, data, and weights are available at \url{https://github.com/mvrl/TerraDiT}.
\end{abstract}

\section{Introduction}
\label{sec:intro}

Controllable generative modeling has advanced rapidly in recent years. Beyond text-to-image synthesis, modern frameworks allow generation to be guided by spatial signals such as depth maps, edge maps, segmentation layouts, or bounding boxes~\cite{mou2023t2iadapter, huang2023composer, zhang2023adding, li2023gligen, wang2024instancediffusion}. Architectures for spatial grounding, including ControlNet~\cite{zhang2023adding} and instance-grounded approaches~\cite{wang2024instancediffusion, li2023gligen}, enable precise spatial control in natural images featuring few, dominant objects. However, these models struggle in the remote sensing (RS) domain, where scenes consist of dense, interconnected arrangements of small elements~\cite{siampoupoly2vec, wang2021loveda}.

To bridge this gap, prior work in geospatial generation adapts existing controllable architectures using geographic data sources such as OpenStreetMap, which provide rich vector annotations describing roads, buildings, land use regions, and other geospatial features~\cite{arsanjani2015introduction}. Such annotations are converted into vision-compatible formats like pixel maps, bounding boxes, or point prompts to guide synthesis~\cite{cher2026vectorsynth, toker2024satsynth, Benidir2025TheCY, tang2025aero, sastry2024geosynth, sastry2026terradit}. The resulting conditional models benefit Earth observation applications such as disaster response, environmental monitoring, and urban planning~\cite{Goktepe2025EcoMapperGM, he2021spatial, mahara2024multispectral, pang2023ssrgan}.

While these adaptations make existing architectures applicable to satellite imagery, they introduce two fundamental limitations. First, converting native geospatial primitive formats degrades their precise geometric properties~\cite{siampoupoly2vec}. For example, single-channel pixel maps ambiguously flatten complex overlaps, rendering a layered overpass indistinguishable from a standard intersection. Vector primitives explicitly preserve this structural continuation, alongside other multi-instance overlaps like land-use zones intersecting buildings. While multi-channel pixel maps can resolve overlaps, they create severe computational bottlenecks in dense satellite scenes with significantly more instances than natural imagery.

Second, format conversion misaligns with practical GeoAI workflows. Dense semantic maps require prohibitive annotation efforts: on average, annotating a single OpenEarthMap segmentation tile takes 2.5 hours~\cite{xia2023openearthmap}. Conversely, while sparse signals like point prompts scale cheaply, they underspecify crucial geometric layouts. Locking into a single supervision granularity forces practitioners to deploy separate architectures for different tasks, making large-scale deployment impractical. A truly robust end-to-end framework should support varying annotation budgets—from cheap point prompts to precise polygons—without requiring format conversion.

To address these limitations, we propose \TerraDiTOmega, a unified framework that directly consumes native geospatial primitives (polygons, polylines, bounding boxes, and points) to guide satellite image synthesis (see Fig.~\ref{fig:teaser}). Directly operating on primitives introduces new technical challenges: geospatial features exhibit diverse geometric complexity and spatial correlations with surrounding instances~\cite{siampoupoly2vec}. While prior grounding frameworks encode spatial controls through generic learnable tokens~\cite{li2023gligen, wang2024instancediffusion}, this \textit{implicit} spatial learning struggles in satellite scenes containing many small, overlapping instances. Furthermore, generic tokens fail to account for the varying degrees of structural detail across different input formats. To address this, we propose a conditioning mechanism that injects \textit{explicit} geometric cues tailored to the complexity of each input primitive.

By directly modeling geospatial primitives, which serve as the foundational basis of remote sensing annotations, our framework is also capable of versatile synthetic data generation. Rather than requiring separate, task-specific generative pipelines, our single unified framework inherently supports a wide variety of downstream tasks. Our contributions are summarized as follows:
\begin{itemize}
    \item We introduce \textbf{\TerraDiTOmega}, a unified generative framework that directly operates on any geospatial primitive (polygons, polylines, boxes, and points), enabling state-of-the-art controllable synthesis across conditioning formats.
    \item We propose \textbf{Geometry-Aware Local Attention}, a conditioning mechanism that \textit{explicitly} preserves geometric cues in vector primitives, improving fidelity in densely structured scenes.
    \item Through our single architecture, we demonstrate higher downstream performance across 4 RS tasks via synthetic data augmentation: \textit{land-cover segmentation, object detection, road graph extraction,} and \textit{scene classification.}
\end{itemize}

\section{Related Works}
\label{sec:related_works}

\subsubsection*{Spatially Controlled Image Generation.}
Spatially controlled image synthesis guides generation using layout-based conditions~\cite{feng2023layoutgpt, zhang2023adding, li2023gligen, wang2024instancediffusion, isola2017image, liu2017unsupervised, vandenOord2016conditional, wang2018high, xu2018attngan, zhang2021ufc, zhu2017unpaired, gafni2022makeascene, avrahami2022spatext, guo2025unimc}. Make-A-Scene~\cite{gafni2022makeascene}, SpaText~\cite{avrahami2022spatext}, and ControlNet~\cite{zhang2023adding} offer finer-grained spatial control, conditioning on inputs such as pixel-level segmentation masks. Moving beyond rasters, recent methods explore discrete conditions: GLIGEN~\cite{li2023gligen} encodes singular formats (e.g., bounding boxes), while InstanceDiffusion~\cite{wang2024instancediffusion} supports mixed inputs (boxes, points, masks, scribbles). \textbf{Discussion.} Despite their success in natural images, these discrete approaches present three major limitations in the RS domain. First, their vision-centric formats stall GeoAI workflows; loading per-instance masks takes $6 \times$ longer than primitives due to the high number of instances. Second, both GLIGEN and InstanceDiffusion use an identical encoding mechanism for different condition types, ignoring the heavily varying complexity in different geospatial primitives. Third, injecting generic tokens via \textit{implicit} attention mechanisms imposes a representational burden in satellite imagery, which contains dense, oriented objects with complex interdependencies. In contrast, our approach directly ingests geospatial primitives, \textit{explicitly} providing varying geometric priors tailored to each format.

\subsubsection*{Satellite Image Synthesis.}
Recent advances in generative modeling for remote sensing largely extend general text-to-image frameworks to satellite imagery~\cite{xu2023txt2img, zhang2024rs5m, sastry2026terradit, liu2025text2earth}. Unlike previous U-Net based architectures, TerraDiT-$\alpha$~\cite{sastry2026terradit} employs a diffusion transformer~\cite{peebles2023scalable} for text-to-image generation in the RS domain. To provide finer-grained control, prior works have explored incorporating metadata (e.g., time or geolocation), rasterized inputs (e.g. segmentation maps or multispectral imagery), or sparse annotations (point prompts)~\cite{khanna2023diffusionsat, tang2024crs, sastry2024geosynth, cher2026vectorsynth, sastry2026terradit}. Notably, to lower annotation cost, TerraDiT-$\Sigma$~\cite{sastry2026terradit} sparsely samples point prompts to provide structural guidance. These contextual approaches have demonstrated promising results across downstream applications such as cloud removal~\cite{wang2023cloud}, urban planning~\cite{wang2025urban}, temporal prediction~\cite{khanna2023diffusionsat}, and super-resolution~\cite{he2021spatial}. However, prior methods rely on converted representations (e.g., pixel maps) and a single layout format. In contrast, we directly use native geospatial primitives, supporting multiple formats and aligning naturally with GeoAI workflows.

\subsubsection*{Remote Sensing Synthetic Data Augmentation.}
Unlike natural imagery readily collected at internet scale~\cite{wang2024skyscript}, much remote sensing data remains unlabeled because annotation requires specialized expertise~\cite{wang2023samrs}. The process is further inefficient since objects are sparse and small, often taking hours to annotate a single tile~\cite{xia2023openearthmap}. To this end, prior works have explored generative models for synthetic data augmentation to improve downstream remote sensing tasks. SatSynth~\cite{toker2024satsynth} and AeroGen~\cite{tang2025aero} train conditional diffusion models to improve downstream model performance on land-cover segmentation and object detection, respectively. Other works~\cite{benidir2025change, zhang2023diffucd, song2023syntheworld} have explored using masks and maps to improve change detection downstream performance. However, these methods require a dedicated architecture for each remote sensing task. Instead, by directly conditioning on all native geospatial primitives, our unified approach can generate synthetic imagery for \textit{multiple} downstream tasks.

\section{Methodology}
\label{sec:method}

In this section, we formalize our generative approach (\ref{subsec:prelims}) and dataset construction (\ref{subsec:dataset}). We then introduce our conditioning strategy (\ref{subsec:core_method}), consisting of a Unified Primitive Encoder to encode geospatial primitives and Geometry-Aware Local Attention (GALA) to effectively condition the model on geometric cues.

\subsection{Preliminaries}
\label{subsec:prelims}
We follow the flow-based generative modeling framework~\cite{lipman2022flow, liu2022flow, albergo2023stochastic}, where a model learns a continuous transformation that transports Gaussian noise $\epsilon \sim \mathcal{N}(0,\mathbf{I})$ to the data distribution $\hat{\mathbf{x}} \sim p(\mathbf{x})$ by predicting a time-dependent velocity field $\mathbf{v}(\mathbf{x}_t,t)$. Noisy states along the flow are written as:
\begin{equation}
    \mathbf{x}_t = \alpha_t \hat{\mathbf{x}} + \sigma_t \epsilon,
\end{equation}
where $\alpha_t$ and $\sigma_t$ are drift and diffusion schedules. The evolution of $\mathbf{x}_t$ is governed by the probability-flow ODE
\begin{equation}
    \dot{\mathbf{X}}_t = \mathbf{v}(\mathbf{x}_t,t),
\end{equation}
and samples are obtained by integrating this ODE from $t{=}1$ (noise) to $t{=}0$ (data). The velocity field is parameterized by a neural network $\mathbf{v}_\theta$ trained using
\begin{equation}
    \mathcal{L}_\mathbf{v}
    =
    \mathbb{E}_{\hat{\mathbf{x}},\epsilon,t}
    \left[
    \left\|
    \mathbf{v}_\theta(\mathbf{x}_t,t)
    -
    \dot{\alpha}_t \hat{\mathbf{x}}
    -
    \dot{\sigma}_t \epsilon
    \right\|^2
    \right].
\end{equation}

\subsubsection*{TerraDiT.}
Building on the flow-based framework, TerraDiT~\cite{sastry2026terradit} introduces text conditioning via cross-attention (TerraDiT-$\alpha$), and geolocation and point prompt conditioning (TerraDiT-$\Sigma$) via Adaptive Local Attention (ALA). ALA guides cross-attention using axis-aligned Gaussian priors centered on point prompts. While effective for sparse points, ALA remains limited: (i) its axis-aligned priors cannot model rotated or elongated geometry; (ii) it does not extend to richer formats such as bounding boxes, polygons, or polylines; and (iii) it lacks explicit instance-level correspondence. We address these limitations with an instance-level framework capable of adapting to any geospatial primitive.

\begin{figure}[ht]
    \centering
    \includegraphics[width=0.8\linewidth]{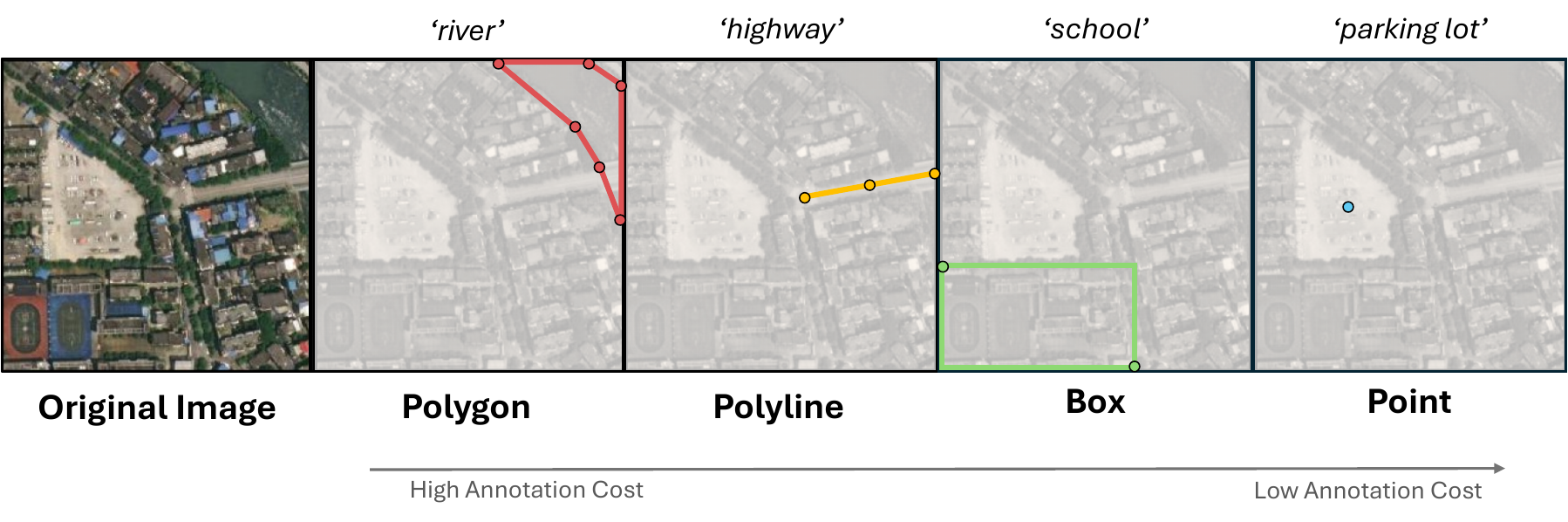}
    \caption{\textbf{Geospatial Primitive Conditions.} Examples of native geospatial primitives $\{\text{polygon, polyline, bounding box, point}\}$. Polygons and polylines are represented as coordinate sequences, bounding boxes as corner pairs, and points as single coordinates.}
    \label{fig:dataset}
\end{figure}

\subsection{Dataset Construction}
\label{subsec:dataset}
We construct a dataset of aligned geospatial primitives, satellite imagery, geolocation, and instance captions following~\cite{liu2025text2earth, cher2026vectorsynth, sastry2026terradit}. Satellite images, global text, and geolocations are sourced from Git-10M~\cite{liu2025text2earth}. For each image tile, we retrieve the corresponding OpenStreetMap (OSM) vector geometry data via Mapbox, using their associated semantic tags as instance captions. Fig.~\ref{fig:dataset} visualizes the set of polygon, polyline, bounding box, and point conditions. See further dataset details in the Appendix.

\subsection{\TerraDiTOmega}
\label{subsec:core_method}
\begin{figure}[ht]
    \centering
    \includegraphics[width=0.9\linewidth]{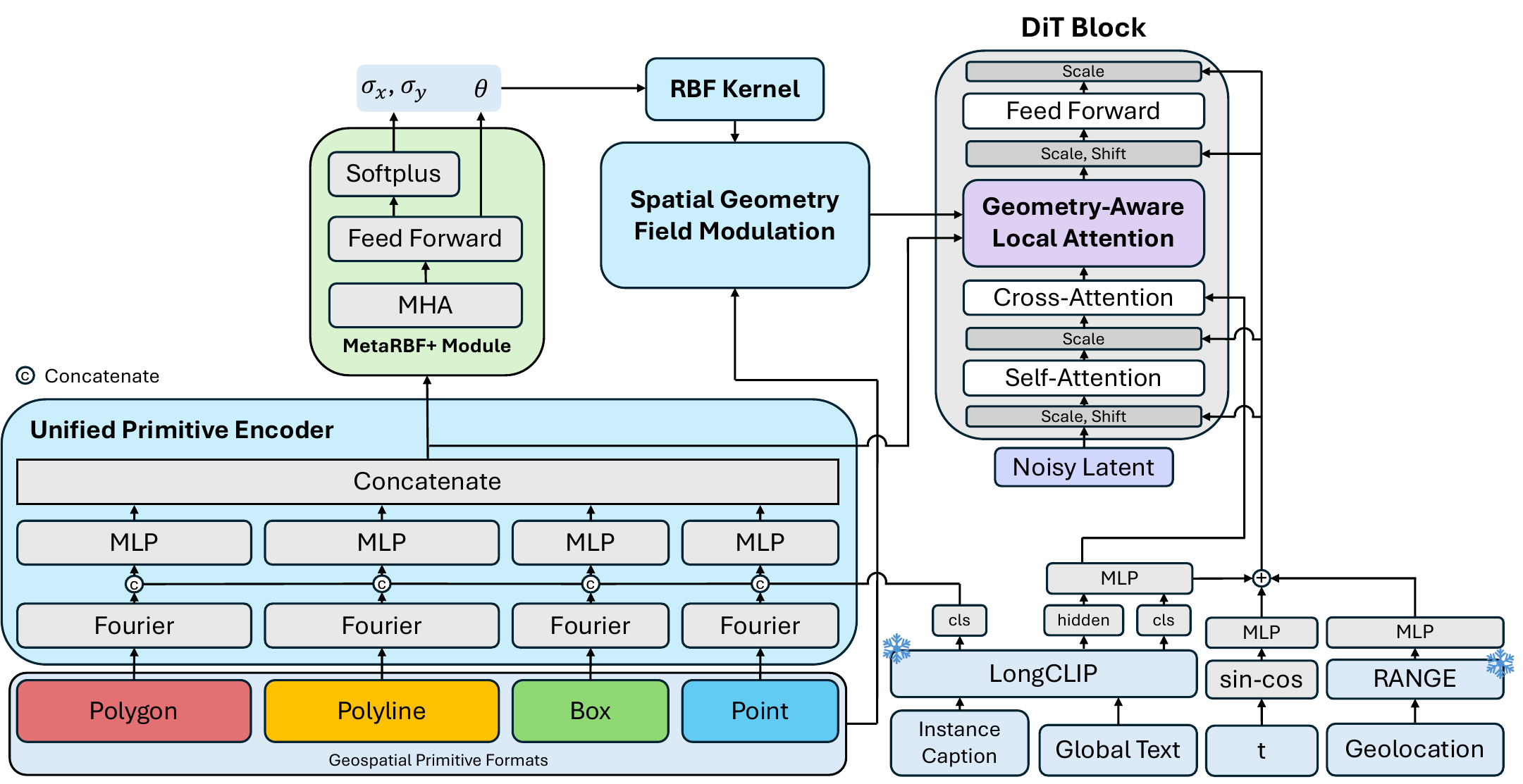}
    \caption{\textbf{\TerraDiTOmega{} Architecture.} A Unified Primitive Encoder encodes primitives for MetaRBF+, which predicts Gaussian kernel parameters ($\sigma_x, \sigma_y, \theta$), modulated by a Spatial Geometry Field. The resulting signal regularizes attention between primitive and visual tokens through Geometry-Aware Local Attention.}
    \label{fig:omega_architecture}
\end{figure}

We employ a Latent Diffusion Transformer~\cite{peebles2023scalable} as the backbone of \TerraDiTOmega. Global text is incorporated through cross-attention, while geolocation conditioning is applied via adaptive layernorm. As shown in Fig.~\ref{fig:omega_architecture}, our core contribution is native geospatial primitive conditioning (polygons, polylines, bounding boxes, points). A Unified Primitive Encoder maps these formats into a shared representation, while Geometry-Aware Local Attention (GALA) injects spatial priors. Specifically, GALA uses MetaRBF+ for coarse point regularization and Spatial Geometry Field modulation for finer, geometry-aligned priors in more complex primitives.

\subsubsection*{Unified Primitive Encoder.}
We employ a unified encoder that maps polygons, polylines, bounding boxes, and points into a shared representation. Each instance may consist of any subset of these primitives and a single associated instance caption. For an instance $i$, we denote the given geospatial primitives $\{\mathbf{p}^{\text{poly}}_i,\, \mathbf{p}^{\text{line}}_i,\, \mathbf{p}^{\text{box}}_i,\, \mathbf{p}^{\text{pt}}_i\}$, where $\mathbf{p}^{\text{poly}}_i \in \mathbb{R}^{V_p \times 2}$ is a polygon with at most $V_p$ vertices, $\mathbf{p}^{\text{line}}_i \in \mathbb{R}^{V_\ell \times 2}$ is a polyline with at most $V_\ell$ vertices, $\mathbf{p}^{\text{box}}_i=(x^{\text{tl}},y^{\text{tl}},x^{\text{br}},y^{\text{br}})$ is an axis-aligned bounding box,
and $\mathbf{p}^{\text{pt}}_i$ is a point. Each instance caption $t_i$ is encoded by a frozen LongCLIP~\cite{zhang2024long} model into embedding $\mathbf{e}_i\in\mathbb{R}^{d_t}$.

Each primitive is encoded with a Fourier feature mapping $\gamma(\cdot)$, concatenated with the text embedding $\mathbf{e}_i$, and passed through a format-specific MLP to produce primitive features:
$\mathbf{g}^{\text{poly}}_i$,
$\mathbf{g}^{\text{line}}_i$,
$\mathbf{g}^{\text{box}}_i$, and
$\mathbf{g}^{\text{pt}}_i$.
When a primitive is absent, it is represented by a learned, primitive-specific null embedding. Finally, we concatenate features into a unified representation $\mathbf{g}_i =\big[\mathbf{g}^{\text{poly}}_i,\, \mathbf{g}^{\text{line}}_i,\, \mathbf{g}^{\text{box}}_i,\, \mathbf{g}^{\text{pt}}_i\big].$

\begin{figure}[ht]
    \centering
    \includegraphics[width=0.66\linewidth]{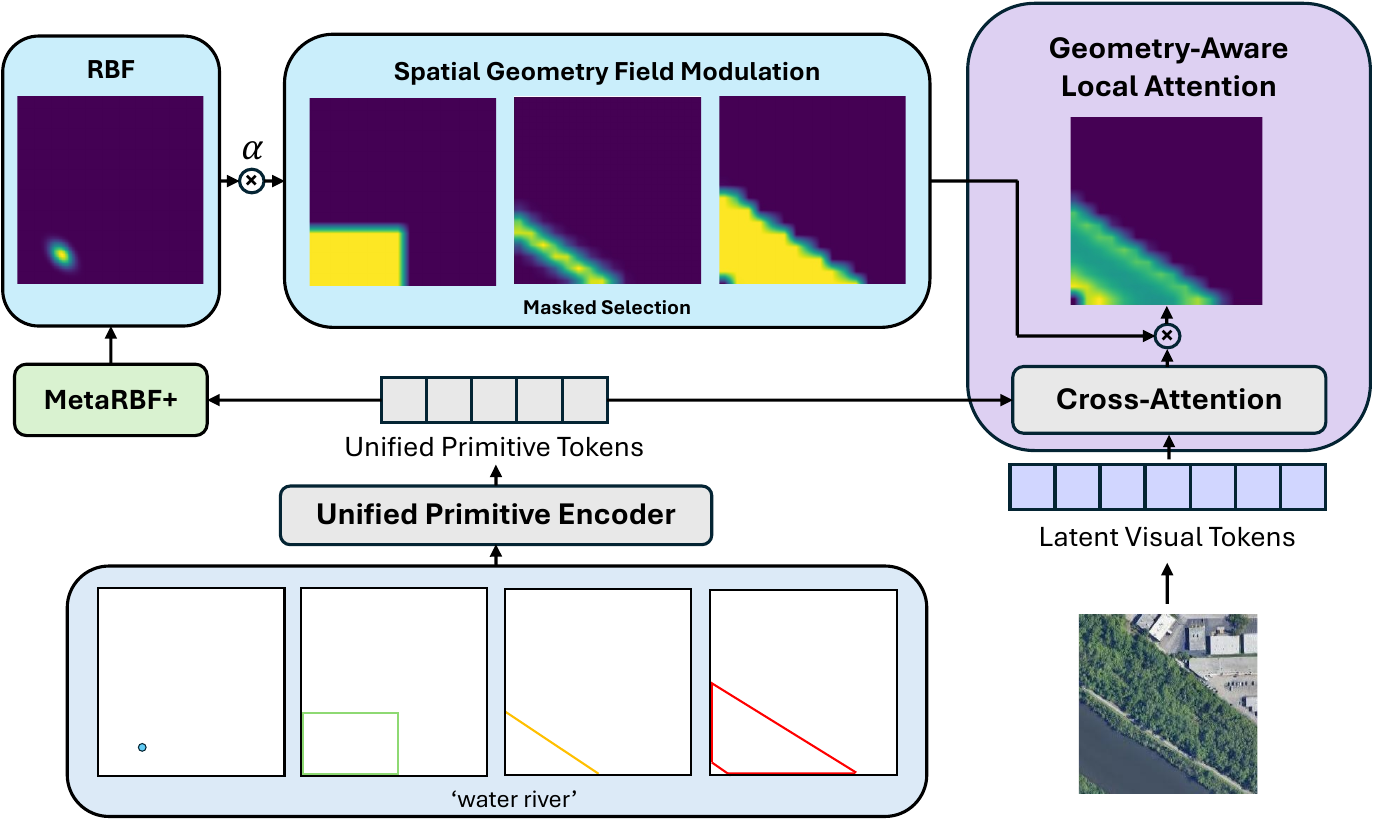}
    \caption{\textbf{Geometry-Aware Local Attention (GALA).} Unified primitive tokens cross-attend with visual tokens, which are then multiplicatively modulated by a Gaussian prior and, for complex primitives, a spatial geometry field.}
    \label{fig:gala}
\end{figure}

\subsubsection*{Geometry-Aware Local Attention.}
As seen in Fig.~\ref{fig:gala}, we ground unified primitive tokens $\mathbf{g}$ using Geometry-Aware Local Attention (GALA). Building on the axis-aligned RBFs of TerraDiT-$\Sigma$'s Adaptive Local Attention, GALA learns a \emph{rotated} anisotropic Gaussian prior to adapt to oriented and elongated geometries. For each instance token $\mathbf{g}_i$, we predict Gaussian kernel parameters  $(\sigma_x, \sigma_y, \theta)$ using a MetaRBF+ module, consisting of multi-head attention (MHA) followed by a feed forward MLP. We apply softplus to $\sigma_x$ and $\sigma_y$ to enforce positivity, leaving the rotation angle $\theta$ unconstrained. The rotated anisotropic Gaussian prior over pixel coordinates $(x,y)$ is described as:
\begin{equation}
K_{\text{pt}}(x,y \mid g_i)
=
\exp\!\Bigg(
-\tfrac{1}{2}
\left\|
\begin{bmatrix}
\sigma_x^{-1} & 0 \\
0 & \sigma_y^{-1}
\end{bmatrix}
\begin{bmatrix}
\cos\theta & -\sin\theta \\
\sin\theta & \cos\theta
\end{bmatrix}
\begin{bmatrix}
x - x_i \\
y - y_i
\end{bmatrix}
\right\|_2^2
\Bigg),
\end{equation}
where $(x_i,y_i)$ denotes the center coordinate of the instance. Compared to the axis-aligned kernel in ALA~\cite{sastry2026terradit}, this formulation allows the attention prior to stretch and rotate to better align with instance orientation.

While point-based priors provide soft localization, richer geometric conditions (polygons, polylines, or bounding boxes) require stronger spatial constraints. We therefore introduce a \emph{spatial geometry field} $S(x,y)$ that explicitly injects geometric cues as a dense multiplicative modulation over the image lattice. For polygons and bounding boxes, we compute a signed distance field (SDF) with respect to their boundaries and apply a sigmoid falloff $\sigma(\cdot)$,
\begin{equation}
S^k(x,y) = \sigma\!\big(-\alpha_k \, \mathrm{SDF}_k(x,y)\big), \quad k \in \{\text{poly}, \text{box}\}
\label{eq:poly_box}
\end{equation}
where $\alpha_{\text{poly}}$ and $\alpha_{\text{box}}$ are learnable strength parameters constrained to be positive via softplus. Negative SDF values correspond to points inside the region, producing high responses near the interior and sharp decay outside.
For polylines, we compute the distance to the nearest line segment and apply a contrast function $h(\cdot)$, effectively expanding the line support since polylines do not capture the full visual thickness of structures such as rivers (see Fig.~\ref{fig:gala}):
\begin{equation}
S^{\text{line}}(x,y)
=
\exp\!\big(-\alpha_{\text{line}} \, h(d_{\text{line}}(x,y))\big),
\end{equation}
where $d_{\text{line}}$ denotes the normalized distance to the polyline and $h(d)=d^2$ in our implementation. This yields a narrow tubular regularization aligned with the polyline geometry, as seen in Fig.~\ref{fig:gala}. More complex primitives define a spatial field, and a format mask selects the active field, prioritizing polygons and polylines over bounding boxes as they provide finer spatial precision. The selected field forms the final spatial geometry field $S_\text{final}(x, y)$. We apply format-wise dropout during training to encourage robustness to varying geometric supervision. Finally, we form the normalized geometry-aware prior:
\begin{equation}
K_{\text{final}}(x,y \mid g_i)
=
\frac{K_{\text{pt}}(x,y \mid g_i)\, S_{\text{final}}(x,y)}
{\sum_{x',y'} K_{\text{pt}}(x',y' \mid g_i)\, S_{\text{final}}(x',y')},
\end{equation}
which multiplicatively modulates the attention logits between image and primitive tokens. As a result, GALA yields format-adaptive grounding: points induce soft local focus, while polylines, polygons, and boxes progressively sharpen attention to respect the full instance geometry.

\section{Implementation Details}
\label{sec:implementation}
\subsubsection*{Training Details.}
We initialize \TerraDiTOmega from TerraDiT-$\alpha$ and fine-tune for 200k steps with AdamW~\cite{loshchilov2017decoupled_adamw} at a learning rate of \texttt{1e-5}. We maintain a total batch size of 256 across 4 NVIDIA H100 GPUs. Consistent with the empirical findings of TerraDiT~\cite{sastry2026terradit}, we perform representation alignment (REPA)~\cite{yu2024representation} using the satellite-specific DINOv3 encoder~\cite{simeoni2025dinov3}, incorporate geolocation conditioning via RANGE~\cite{dhakal2025range}, and encode global text with a frozen LongCLIP~\cite{zhang2024long} encoder. For ablation studies, all models are trained from scratch using the base (B) variant for 400k steps.  Additional details are provided in the Appendix.

\begin{figure*}[t!]
    \centering
    \includegraphics[width=0.9\linewidth]{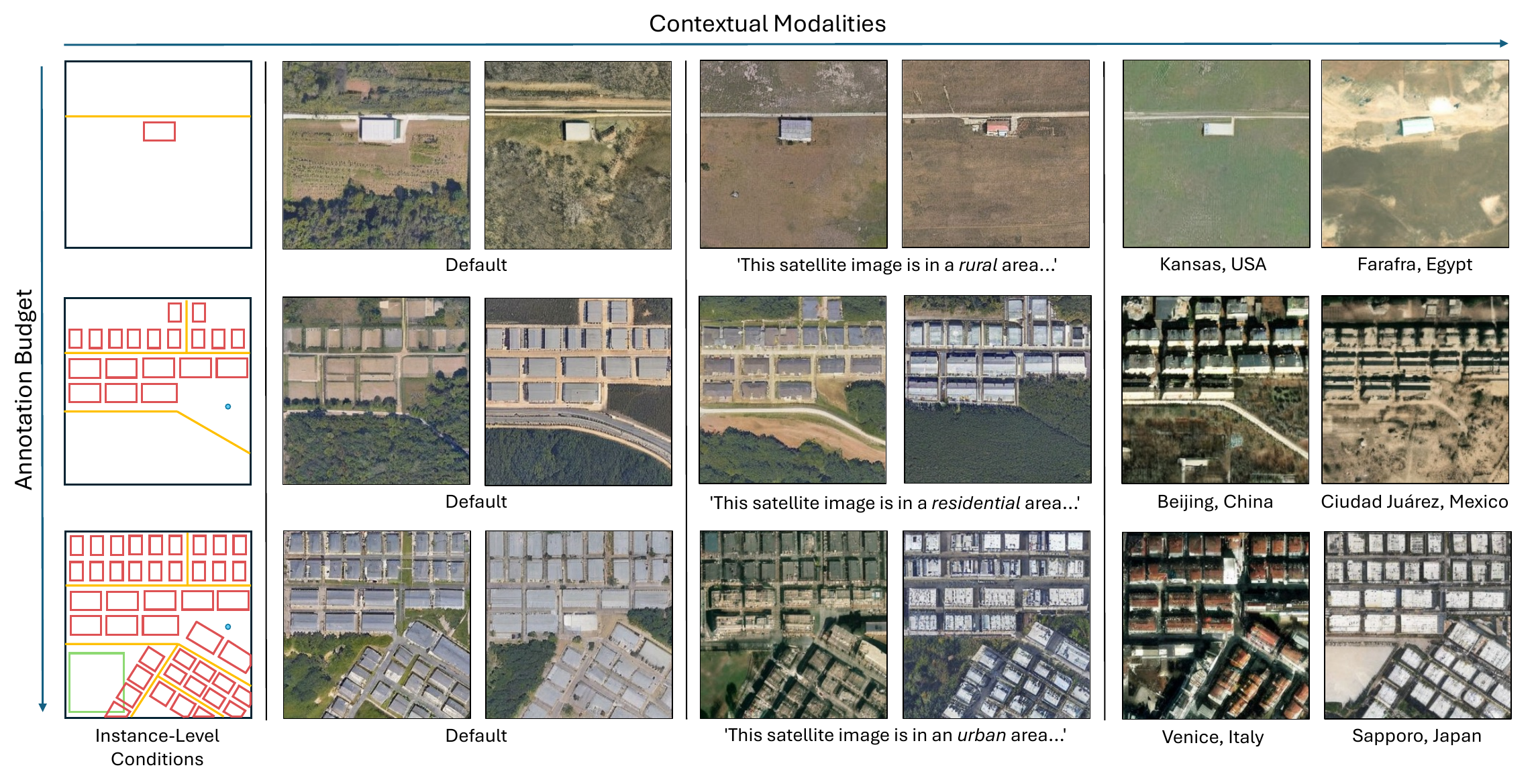}
    \caption{\textbf{Qualitative Samples from TerraDiT-\TextOmega.} Conditions increase in complexity and annotation budget from top to bottom, while contextual modalities (global text and geolocation) are progressively added from left to right. \textcolor{polyRed}{Polygons} denote `building', \textcolor{lineYellow}{polylines} denote `road', \textcolor{boxGreen}{bounding boxes} denote `recreational', and \textcolor{pointBlue}{points} denote `trees'.}
    \label{fig:qual_omega}
\end{figure*}

\section{Experiments}
\label{sec:exp}

To assess synthesis quality and condition alignment, we evaluate over a suite of metrics. We use FID~\cite{heusel2017gans}, sFID~\cite{nash2021generating}, LPIPS~\cite{zhang2018perceptual}, and CLIP Score~\cite{hessel-etal-2021-clipscore} to evaluate visual fidelity, spatial fidelity, perceptual realism, and text-image alignment, respectively. To evaluate grounding accuracy with spatial conditions, we use Classification Accuracy Score (CAS)~\cite{ravuri2019classification}. In addition, we use SSIM to assess primitive structural adherence. Finally, we measure downstream data augmentation utility across four practical tasks: bounding box precision for object detection (mAP@50-95, mAP@50), pixel-level overlap for segmentation (mIoU, F1), topological connectivity for road graph extraction (TOPO~\cite{biagioni2012inferring}, APLS~\cite{vanetten2018spacenet}), and accuracy for scene classification (Acc-1, Acc-5). See further metric details in the Appendix.

\begin{figure*}[t!]
    \centering

    \includegraphics[width=0.9\linewidth]{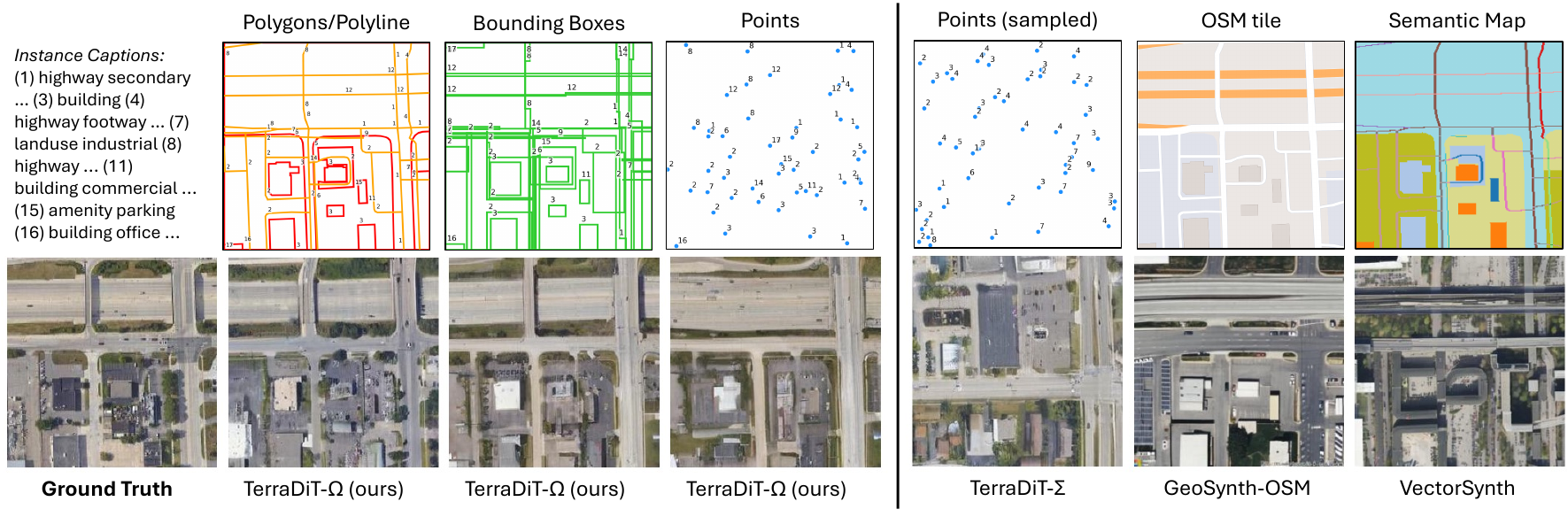}
    \caption{Qualitative comparison with geospatial grounding models.}
    \label{fig:qual_comparison}

    \par\medskip

    \begin{minipage}[t]{0.6\linewidth}
        \hbox{\vtop{\vbox{\hsize=\linewidth\centering\includegraphics[width=0.95\linewidth]{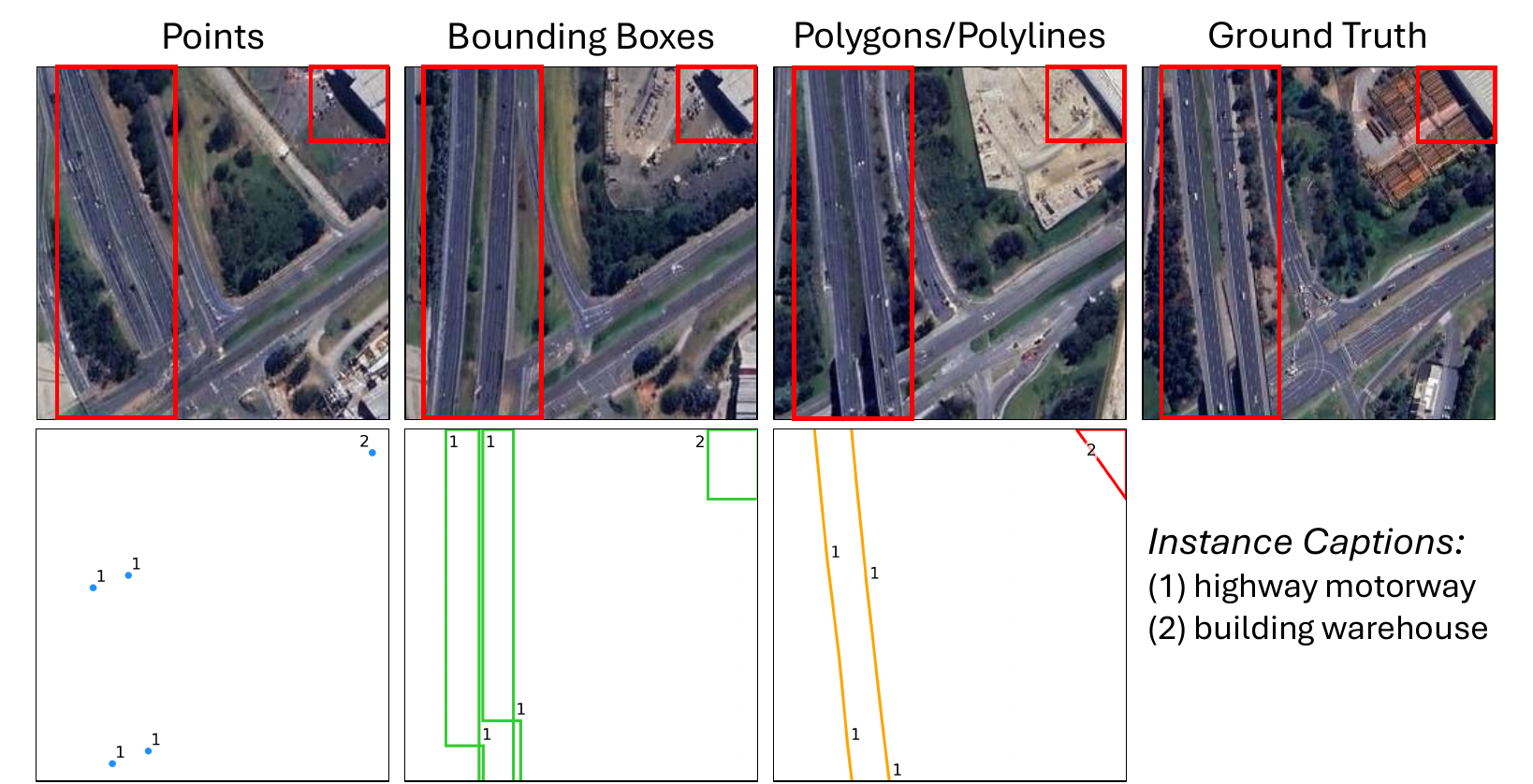}}}}
        \caption{\textbf{Comparison of Formats.} Spatial alignment improves with richer conditioning. \textcolor{lineYellow}{Polylines} better preserve the highway direction (left) and \textcolor{polyRed}{polygons} align more tightly with the warehouse building (top right). Imagery is best viewed zoomed in.}
        \label{fig:point_box_poly}
    \end{minipage}
    \hfill
    \begin{minipage}[t]{0.365\linewidth}
        \hbox{\vtop{\vbox{\hsize=\linewidth\centering\includegraphics[width=\linewidth]{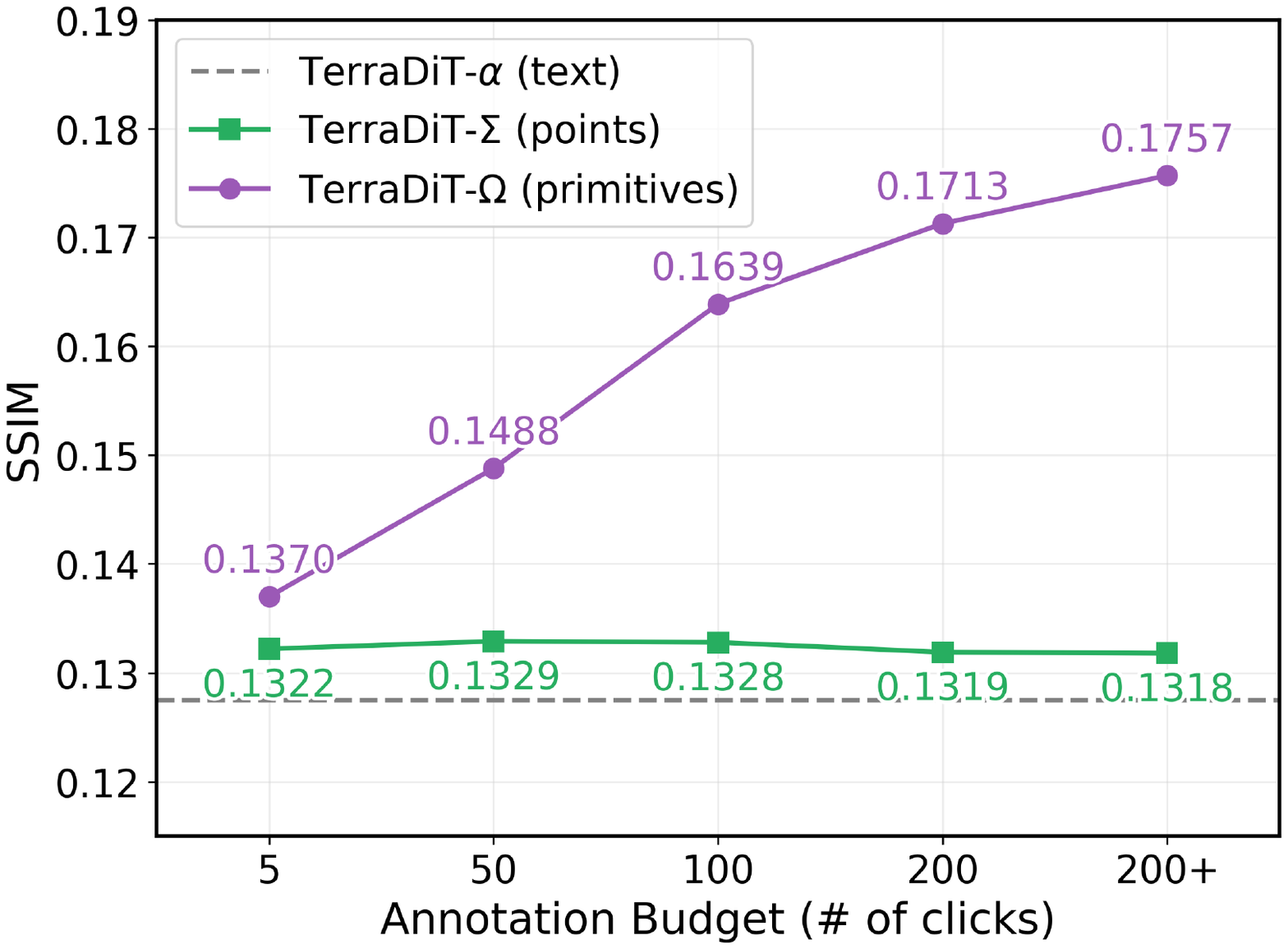}}}}
        \caption{Structural similarity with real imagery improves as the annotation budget scales. SSIM ($\uparrow$) is evaluated on Git-Dense-3.5k.}
        \label{fig:annotation_scaling}
    \end{minipage}
\end{figure*}

\subsection{Qualitative Evaluations}
Fig.~\ref{fig:qual_omega} demonstrates that \TerraDiTOmega faithfully adheres to spatial layouts across varying primitive densities while seamlessly integrating global text and geolocation for precise stylistic control (e.g., generating taller structures for urban prompts and sandy textures for Egypt and Mexico). Furthermore, our primitive-level conditioning preserves structural fidelity significantly better than raster-based baselines like GeoSynth-OSM~\cite{sastry2024geosynth} and VectorSynth~\cite{cher2026vectorsynth} (Fig.~\ref{fig:qual_comparison}). Raster methods struggle with overlapping geometries—such as incorrectly ordering intersecting overpasses—whereas our polyline encoding explicitly maintains correct relational connectivity. Compared to other coordinate-based methods, our instance-grounded points offer finer geometric accuracy than the point-based sampling in TerraDiT-$\Sigma$~\cite{sastry2026terradit}. In addition, Fig.~\ref{fig:point_box_poly} illustrates that richer formats (polygons/polylines) reduce layout hallucinations by providing exact geometric and orientation details. For example, polylines better preserve the exact trajectory of highways, and polygons enable tighter alignment with building edges compared to points or bounding boxes.

\begin{table*}[!t]
\centering
\caption{\textbf{Zero-shot evaluation on Git-Rand-15k, Git-Spatial-15k, and Git-Dense-3.5k.}
$\downarrow$ indicates lower is better, while $\uparrow$ indicates higher is better.
$T$ is text input, $O$ is OSM raster input, $L$ is geolocation, $P$ is points, $B$ is bounding box, and $\Omega$ is \{polygon, polyline, bounding box, and point conditioning\}.
\scriptsize{\textit{*Text2Earth was trained on the entire Git-10M dataset (including our test sets), which may inflate reported performance.}}}
\label{tab:main_quant}
\setlength{\aboverulesep}{0pt}
\setlength{\belowrulesep}{0pt}
\resizebox{\linewidth}{!}{
\begin{tabular}{l|c|ccc|ccc|ccc}
\multirow{2}{*}{Model} & \multirow{2}{*}{Condition} &
\multicolumn{3}{c|}{\textbf{Git-Rand-15k}} &
\multicolumn{3}{c|}{\textbf{Git-Spatial-15k}} &
\multicolumn{3}{c}{\textbf{Git-Dense-3.5k}} \\
\cline{3-11}
& & FID $\downarrow$ & sFID $\downarrow$ & LPIPS $\downarrow$ &
FID $\downarrow$ & sFID $\downarrow$ & LPIPS $\downarrow$ &
FID $\downarrow$ & sFID $\downarrow$ & LPIPS $\downarrow$ \\
\toprule
\rowcolor{gray!25}
\multicolumn{11}{l}{\textit{\textbf{General} Models}} \\
SDXL~\cite{podell2023sdxl} & $T$ &
150.10 & 21.27 & 0.5929 &
221.76 & 32.06 & 0.5845 &
187.45 & 27.03 & 0.6076 \\
SD 3~\cite{esser2024scaling} & $T$ &
90.45 & 14.21 & 0.5083 &
149.34 & 23.37 & 0.5353 &
113.16 & 23.43 & 0.5752 \\
PixArt-$\alpha$-XL~\cite{chen2023pixart} & $T$ &
85.82 & 16.30 & 0.5220 &
158.98 & 20.13 & 0.5261 &
91.22 & 25.72 & 0.5463 \\
PixArt-$\Sigma$-XL~\cite{chen2024pixart} & $T$ &
79.36 & 17.67 & 0.4954 &
135.27 & 15.77 & 0.5435 &
84.18 & 27.13 & 0.5515 \\
InstanceDiffusion~\cite{wang2024instancediffusion} & $T + B$ &
94.18 & 22.60 & 0.5975 &
149.42 & 26.60 & 0.6141 &
112.35 & 29.93 & 0.5979 \\
GLIGEN~\cite{li2023gligen} & $T + B$ &
54.00 & 10.82 & 0.4792 &
84.74 & 7.35 & 0.4610 &
96.28 & 24.94 & 0.5256 \\
\midrule
\rowcolor{gray!25}
\multicolumn{11}{l}{\textit{\textbf{Geospatial} Models}} \\
MHN-VQGAN~\cite{xu2023txt2img} & $T$ &
171.37 & 26.56 & 0.4992 &
270.86 & 37.57 & 0.5102 &
93.92 & 52.62 & 0.4869 \\
MHN-VQVAE~\cite{xu2023txt2img} & $T$ &
188.46 & 43.43 & 0.5878 &
204.13 & 42.00 & 0.5695 &
224.27 & 28.00 & 0.6084 \\
GeoRSSD~\cite{zhang2024rs5m} & $T$ &
129.23 & 34.39 & 0.5378 &
206.74 & 45.02 & 0.5564 &
127.95 & 29.58 & 0.5527 \\
CRS-Diff~\cite{tang2024crs} & $T$ &
58.13 & 6.62 & 0.4758 &
93.13 & 7.69 & 0.4644 &
89.98 & 17.54 & 0.5141 \\
DiffusionSat~\cite{khanna2023diffusionsat} & $T+L$ &
71.55 & 19.22 & 0.5449 &
164.79 & 50.02 & 0.5540 &
97.04 & 25.51 & 0.5585 \\
GeoSynth~\cite{sastry2024geosynth} & $T$ &
45.59 & 18.88 & 0.5413 &
85.80 & 31.53 & 0.4384 &
83.90 & 22.26 & 0.5149 \\
GeoSynth-OSM~\cite{sastry2024geosynth} & $T+O$ &
53.88 & 18.69 & 0.4298 &
110.72 & 30.30 & 0.4374 &
82.32 & 20.78 & 0.4400 \\
VectorSynth~\cite{cher2026vectorsynth} & $T+O$ &
95.15 & 35.49 & 0.4441 &
204.16 & 72.35 & 0.4386 &
72.26 & 25.87 & 0.4505 \\
AeroGen~\cite{tang2025aero} & $T$ + $B$ &
85.96 & 14.96 & 0.5225 &
122.86 & 23.91 & 0.4956 &
132.84 & 27.45 & 0.5333 \\
Text2Earth$^*$~\cite{liu2025text2earth} & $T$ &
25.93 & 5.09 & 0.4269 &
49.91 & 5.33 & 0.3926 &
40.23 & 17.37 & 0.4871 \\
TerraDiT-$\alpha$-XL~\cite{sastry2026terradit} & $T$ &
14.21 & 5.13 & 0.3972 &
20.33 & \textbf{4.10} & 0.3863 &
29.76 & 17.94 & 0.4853 \\
TerraDiT-$\Sigma$-XL~\cite{sastry2026terradit} & $T+P+L$ &
12.01 & 5.09 & 0.3779 &
19.18 & 11.36 & 0.3652 &
27.49 & 17.95 & 0.4692 \\
\midrule
& $T+P+L$&
9.91 & 4.63 & 0.3612 &
17.83 & 4.22 & 0.3632 &
20.57 & 17.44 & 0.3876 \\
\textbf{TerraDiT-\TextOmega-XL (ours)} & $T+B+L$ &
\underline{9.72} & \underline{4.45} & \underline{0.3509} &
\textbf{17.46} & 4.16 & \underline{0.3620} &
\underline{19.46} & \underline{16.28} & \underline{0.3594} \\
& $T+\Omega+L$ &
\textbf{9.25} & \textbf{4.38} & \textbf{0.3438} &
\underline{17.54} & \underline{4.15} & \textbf{0.3614} &
\textbf{18.20} & \textbf{16.21} & \textbf{0.3455} \\
\bottomrule
\end{tabular}
}
\end{table*}

\subsection{Quantitative Evaluations}
We quantitatively evaluate \TerraDiTOmega{} against both general image generative models and prior geospatial generative approaches on three held-out in-domain datasets: Git-Rand-15k (randomly sampled across the globe), Git-Spatial-15k (spatial holdout during training), and Git-Dense-3.5k (a subset of Git-Rand-15k restricted to tiles with at least 15 instances). Table~\ref{tab:main_quant} demonstrates that general vision models struggle significantly on our test sets, exhibiting poor realism metrics, confirming that generalist models do not transfer well to the satellite imagery domain. On Git-Rand-15k, \TerraDiTOmega's performance in FID, sFID, and LPIPS consistently scales as the conditioning format progresses from points $\rightarrow$ bounding boxes $\rightarrow$ combined primitives ($T + \Omega + L$). Notably, our point-conditioned model ($T + P + L$) outperforms the TerraDiT-$\Sigma$ point baseline. We attribute this to two factors: 1) our points are grounded to specific instances rather than randomly sampled, and 2) our Gaussian prior encodes rotation, which better preserves the oriented and elongated structures of instances. On the sparser Git-Spatial-15k split, our model achieves state-of-the-art FID and LPIPS. While trailing TerraDiT-$\alpha$ in sFID on this split, our three conditioning schemas still rank second through fourth, demonstrating that our model is robust to sparse spatial conditioning. Finally, on Git-Dense-3.5k, our model achieves its most substantial performance gains over the baselines, highlighting the effectiveness of our approach in challenging, dense environments.

Synthetic data generated by \TerraDiTOmega{} yields substantially higher Classification Accuracy Scores (CAS), as seen in Table~\ref{tab:cas_git}. Notably, CAS scales with primitive complexity: while point-level conditioning already outperforms baselines, adding richer geometric complexity further enhances semantic accuracy. We also evaluate spatial fidelity against annotation budget (measured in clicks) in Fig.~\ref{fig:annotation_scaling}. While the point-based TerraDiT-$\Sigma$ slightly improves upon the text-only TerraDiT-$\alpha$ baseline, it plateaus and cannot benefit from increased annotation effort. In contrast, \TerraDiTOmega{} not only performs better at low budgets, but scales significantly as the budget increases. This scalability stems from our flexible formatting: as the budget grows, \TerraDiTOmega{} effectively translates both the addition of more instances (from 5 up to 15+) and their upgrade to richer primitives into increasingly faithful spatial layouts.

\begin{table*}[t]
    \centering

    \begin{minipage}[t]{0.53\linewidth}
        \centering
        \caption{\textbf{Classification Accuracy Score (CAS) on Git-Rand-15k}. \TerraDiTOmega{} outperforms baseline grounding methods for spatially controlled generation. For CAS, we resize bounding box instances from Git-Rand-15k and train a ResNet-110 on generated imagery to evaluate on real imagery. Upper bound indicates if the generator predicted perfect images.}
        \label{tab:cas_git}
        \smallskip

        \resizebox{\linewidth}{!}{
        \begin{tabular}{l|c|cc}
        Method & Cond & Top-1 $\uparrow $ & Top-5 $\uparrow$ \\
        \toprule
        {\color{gray} Upper Bound} & -- & {\color{gray} 84.23} & {\color{gray} 98.47} \\
        \midrule
        InstanceDiffusion~\cite{wang2024instancediffusion} & box & 54.98 & 88.97 \\
        GLIGEN~\cite{li2023gligen} & box & 65.21 & 92.94 \\
        \midrule
        GeoSynth-OSM~\cite{sastry2024geosynth} & OSM & 64.64 & 90.84 \\
        VectorSynth~\cite{cher2026vectorsynth} & OSM & 61.75 & 92.54 \\
        AeroGen~\cite{tang2025aero} & box & 67.83 & 92.47 \\
        TerraDiT-$\Sigma$-XL~\cite{sastry2026terradit} & points & 70.23 & 95.83 \\
        \midrule
        & points & 76.66 & 97.05 \\
        \textbf{TerraDiT-\TextOmega-XL (ours)} & box & \underline{78.33}& \underline{97.13} \\
        & \TextOmega & \textbf{79.36} & \textbf{97.39} \\
        \bottomrule
        \end{tabular}
        }
    \end{minipage}
    \hfill
    \begin{minipage}[t]{0.44\linewidth}
        \centering
        \caption{\textbf{Ablation Studies.} \textit{(Top)} Impact of adding rotation ($\theta$) via the MetaRBF+ and SGF modulation. \textit{(Middle)} Impact of global text ($T$), geolocation ($L$), and primitives ($\Omega$). \textit{(Bottom)} Choice of attention module for primitive conditioning, using the same Unified Primitive Encoder.}
        \label{tab:ablations_combined}
        \smallskip

        \resizebox{\linewidth}{!}{
        \begin{tabular}{ccc|c|cccc}
        RBF & $\theta$ &  SGF & Iter. & FID $\downarrow$ & sFID $\downarrow$ & SSIM $\uparrow$ & CLIP $\uparrow$ \\
        \midrule
        \xmark & \xmark & \xmark & 400k & 24.08 & 6.65 & 0.2609 & 0.2747 \\
        \cmark & \xmark & \xmark & 400k & 23.52 & 6.58 & 0.2616 & 0.2750 \\
        \cmark & \cmark & \xmark & 400k & 22.41 & 6.41 & 0.2620 & 0.2754 \\
        \rowcolor{gray!25}
        \cmark & \cmark & \cmark & 400k & \textbf{21.97} & \textbf{6.25} & \textbf{0.2632} & \textbf{0.2759} \\
        \toprule
        \end{tabular}
        }

        \par\smallskip

        \resizebox{\linewidth}{!}{
        \begin{tabular}{ccc|c|cccc}
        $T$ & $L$ & $\Omega$ & Iter. & FID $\downarrow$ & sFID $\downarrow$ & SSIM $\uparrow$ & CLIP $\uparrow$ \\
        \midrule
        \xmark & \cmark & \cmark & 400k & 24.90 & 6.72 & 0.2543 & 0.2648 \\
        \cmark & \xmark & \cmark & 400k & 23.53 & 6.59 & 0.2577 & 0.2756 \\
        \cmark & \cmark & \xmark & 400k & 27.19 & 7.34 & 0.2503 & 0.2738 \\
        \rowcolor{gray!25}
        \cmark & \cmark & \cmark & 400k & \textbf{21.97} & \textbf{6.25} & \textbf{0.2632} & \textbf{0.2759} \\
        \toprule
        \end{tabular}
        }

        \par\smallskip

        \resizebox{\linewidth}{!}{
        \begin{tabular}{l|c|cccc}
        Method & Iter. & FID $\downarrow$ & sFID $\downarrow$ & SSIM $\uparrow$ & CLIP $\uparrow$ \\
        \midrule
        Cross Attention & 400k & 24.08 & 6.65 & 0.2609 & 0.2747  \\
        GSA~\cite{li2023gligen} & 400k & 22.27 & 6.32 & 0.2624 & 0.2755 \\
        IMA~\cite{wang2024instancediffusion} & 400k & 22.31 & 6.39 & 0.2627 & 0.2758 \\
        ALA~\cite{sastry2026terradit} & 400k & 23.52 & 6.58 & 0.2616 & 0.2750 \\
        \rowcolor{gray!20}
        \textbf{GALA (ours)} & 400k & \textbf{21.97} & \textbf{6.25} & \textbf{0.2632} & \textbf{0.2759} \\
        \toprule
        \end{tabular}
        }
    \end{minipage}
\end{table*}

\subsection{Ablation Study}
As seen in Table~\ref{tab:ablations_combined}, we perform ablations to analyze the effect of design choices of the proposed geometry-aware conditioning (GALA), the contribution of different input modalities, and compare to previously proposed conditioning methods.

\subsubsection*{Impact of RBF Rotation and SGF Modulation.}
Integrating rotation ($\theta$) into the MetaRBF+ module and adding Spatial Geometry Field (SGF) modulation yields steady improvements across all evaluated metrics. Modeling rotation within the Gaussian prior is crucial for accurately representing oriented and elongated structures, while SGF modulation provides the detailed geometric context necessary for generating complex geospatial primitives.

\subsubsection*{Impact of Input Modalities.}
Evaluating the contribution of global text ($T$), geolocation ($L$), and primitives ($\Omega$) reveals that all three are vital for high-fidelity generation. Notably, the removal of primitive information ($\Omega$) results in the sharpest degradation in performance across all metrics, such as FID increasing from 21.97 to 27.19. This highlights the importance of explicit geospatial primitives in accurately modeling the spatial distribution of satellite imagery.

\begin{figure}[b!]
    \centering
    \includegraphics[width=.9\linewidth]{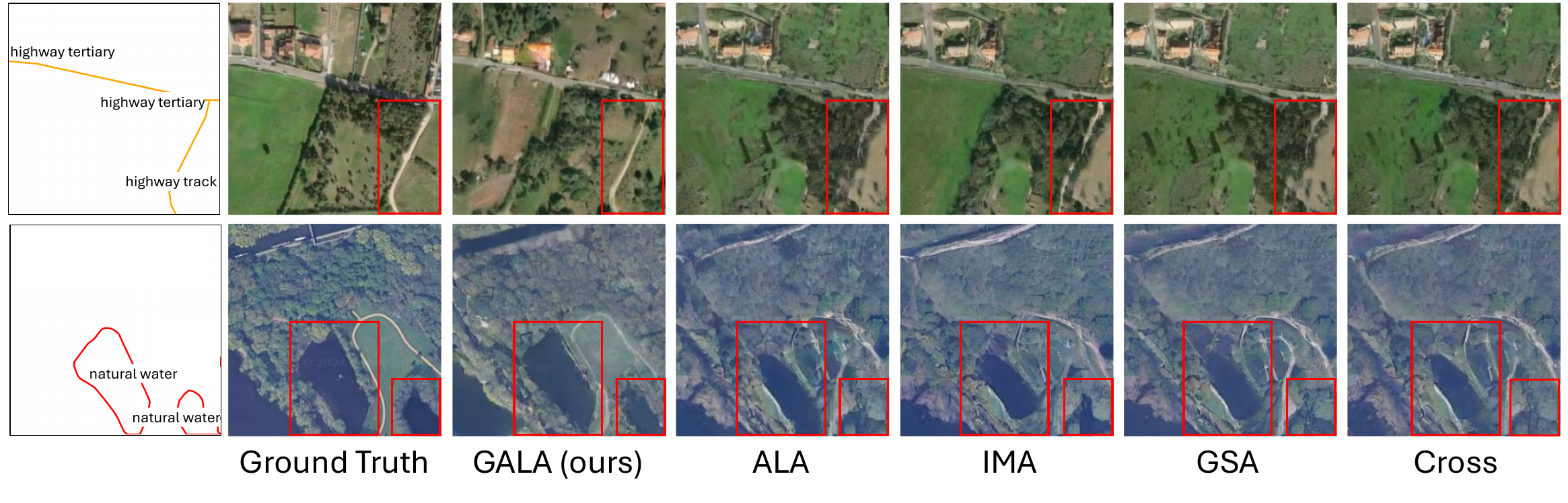}
    \caption{Qualitative Comparison between Geometry-Aware Local Attention (GALA), Adaptive Local Attention (ALA)~\cite{sastry2026terradit}, Instance Masked Attention (IMA)~\cite{wang2024instancediffusion}, Gated Self-Attention (GSA)~\cite{li2023gligen}, and Cross Attention (Cross) after 400k steps. Our approach (GALA), exhibits much stronger structural fidelity compared to other methods.}
    \label{fig:ablate_comparison}
\end{figure}

\subsubsection*{Comparison of Attention Mechanisms.}
To isolate the impact of our GALA module, we evaluate it against vanilla cross-attention and reimplementations of implicit learning methods, including Gated Self-Attention (GSA)~\cite{li2023gligen} and Instance Masked Attention (IMA)~\cite{wang2024instancediffusion}, using the same Unified Primitive Encoder. GALA, which explicitly injects geometric cues into the attention computation, achieves superior performance across all metrics. Furthermore, GALA outperforms other explicit methods like ALA, demonstrating that rigid explicit mechanisms perform worse when they cannot properly adapt to the varying, flexible formats of complex geospatial primitives. Qualitatively, as seen in Fig.~\ref{fig:ablate_comparison}, GALA adheres to complex geometries much more faithfully than competing methods: uniquely capturing the exact curvature of the highway track in the top row and the precise boundaries of both water bodies in the bottom row.

\subsection{Data Augmentation Experiments}
We evaluate the utility of our generated imagery as training data for downstream remote sensing tasks. Specifically, we perform data augmentation for land-cover semantic segmentation (OpenEarthMap~\cite{xia2023openearthmap}), object detection (DIOR~\cite{Li_2020}), road graph extraction (City-Scale~\cite{he2020sat2graph}), and scene classification (AID~\cite{xia2017aid}). After tuning \TerraDiTOmega{} on each task's training set, we generate synthetic imagery by simply applying basic augmentations (flipping, rotation, and transposition) to these source layouts, demonstrating consistent performance gains over real-only baselines.

\begin{figure*}[!ht]
    \centering

    \includegraphics[width=0.95\linewidth]{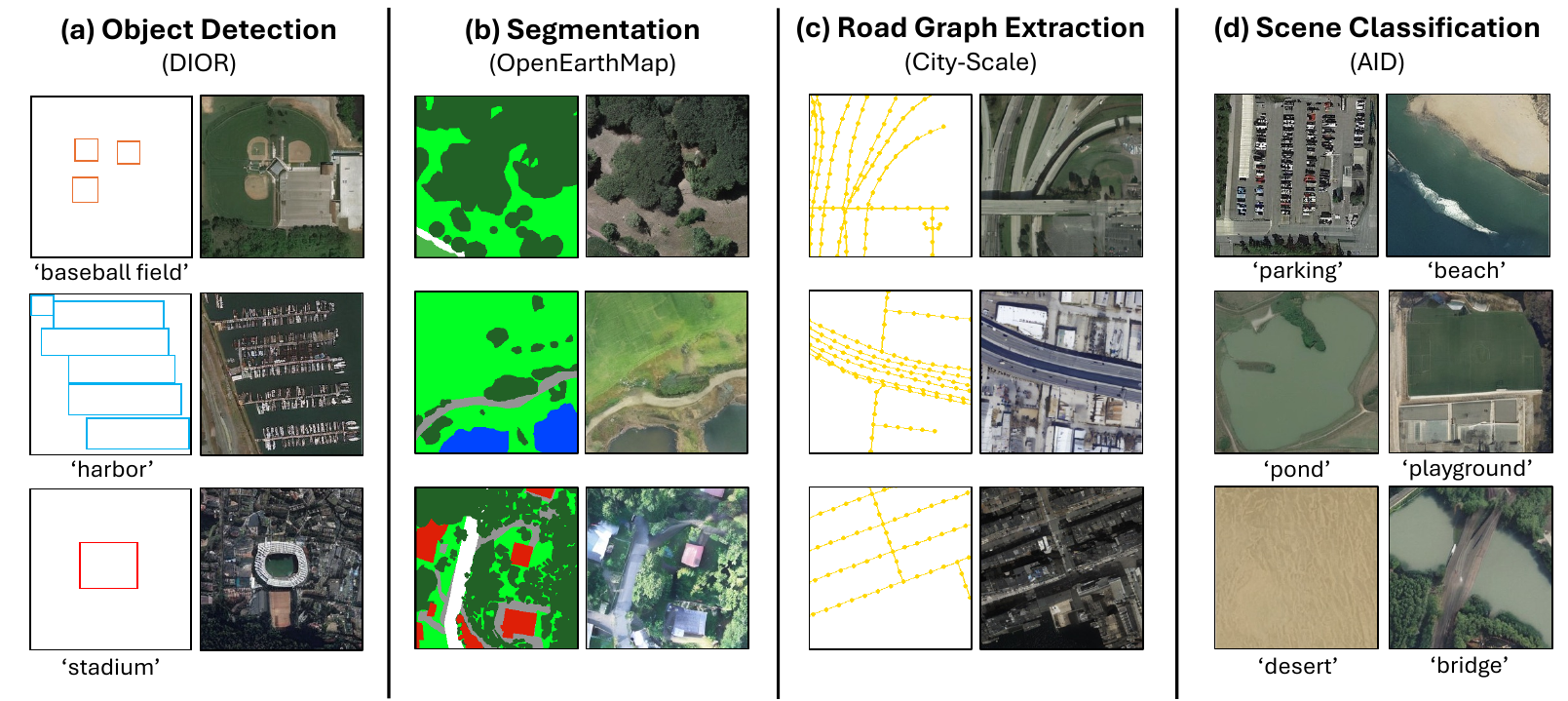}

    \par\medskip

    \resizebox{\linewidth}{!}{
    \begin{tabular}{c|cc|cc|cc|cc}
    \multirow{2}{*}{\makecell{Synth \\ Data}} & \multicolumn{2}{c|}{DIOR} & \multicolumn{2}{c|}{OpenEarthMap} & \multicolumn{2}{c|}{City-Scale} & \multicolumn{2}{c}{AID}\\
    \cline{2-9}
     & mAP@50-95 $\uparrow$ & mAP@50 $\uparrow$ & mIoU $\uparrow$ & F1 $\uparrow$ & TOPO $\uparrow$ & APLS $\uparrow$ & Acc-1 $\uparrow$ & Acc-5 $\uparrow$\\
    \toprule
    $\times0$& 55.15$\pm$0.15 & 77.09$\pm$0.31 & 55.75$\pm$0.26 & 70.70$\pm$0.24 & 73.59$\pm$0.48 & 60.50$\pm$0.60 & 72.67$\pm$0.64 & 94.73$\pm$0.23 \\
    $\times1$& \textbf{56.14$\pm$0.18} & 78.69$\pm$0.15 & 57.05$\pm$0.14 & 71.77$\pm$0.10 & 74.39$\pm$0.10 & 62.58$\pm$0.36 & 81.60$\pm$0.50 & 97.44$\pm$0.22 \\
    $\times2$ & 55.90$\pm$0.17 & \textbf{78.87$\pm$0.28} & \textbf{57.46$\pm$0.12} & \textbf{72.01$\pm$0.12} & \textbf{74.62$\pm$0.17} & \textbf{63.52$\pm$0.32} & \textbf{86.53$\pm$0.38} & \textbf{98.30$\pm$0.17} \\
    \bottomrule
    \end{tabular}
    }

    \smallskip

    \caption{\textbf{Synthetic data augmentation across 4 downstream tasks.} \textit{(Top)} \TerraDiTOmega\ synthetic samples on (a) DIOR, (b) OpenEarthMap, (c) City-Scale, (d) AID. \textit{(Bottom)} We use RT-DETR~\cite{zhao2024detrs} for object detection (DIOR), U-Net~\cite{ronneberger2015u} for land cover segmentation (OpenEarthMap), SAM-Road~\cite{hetang2024segment} for road graph extraction (City-Scale), and a ResNet-34~\cite{he2016deep} for scene classification (AID).}
    \label{fig:synthetic_samples_and_results}
\end{figure*}

\subsubsection{Datasets.}
OpenEarthMap contains high-resolution images across 8 land-cover categories, DIOR includes 20 object classes at varying spatial resolutions, City-Scale has urban satellite imagery spanning 20 U.S. cities, and AID consists of globally diverse images across 30 scene categories. For DIOR and AID, images are resized to $256 \times 256$; for OpenEarthMap and City-Scale, we extract non-overlapping $256 \times 256$ crops from high-resolution tiles, converting OpenEarthMap segmentation masks into polygons. Unlike previous methods~\cite{tang2025aero}, we do not perform any post-processing or filtering on the generated images before training discriminative models, highlighting the consistent spatial fidelity of \TerraDiTOmega. More details for datasets and splits are in the Appendix.

Fig.~\ref{fig:synthetic_samples_and_results} demonstrates synthetic data consistently improves performance across these low-data regimes. On the DIOR object detection task, mAP@50-95 peaks at the $\times 1$ synthetic data ratio (56.14) and drops slightly at $\times 2$ (55.90), whereas mAP@50 continues improving to 78.87 at $\times 2$. We hypothesize that since DIOR is the largest dataset evaluated (18k training samples), the model's overall performance saturates earlier from synthetic augmentation. For OpenEarthMap, the introduction of $\times 1$ and $\times 2$ synthetic data enhances pixel-level grounding, steadily increasing both mIoU and F1 scores. On the City-Scale road extraction task, our synthetic images improve the global TOPO metric ($73.59 \to 74.62$), demonstrating that \TerraDiTOmega{} generates high-quality local road structures that seamlessly connect to form broader global networks. Consequently, APLS, a metric more sensitive to local road geometry, improves more significantly ($+3.02$). We observe dramatic gains in scene classification on the AID dataset, where Top-1 accuracy jumps from $72.67$ without synthetic data to $86.53$ at the $\times 2$ scale. We attribute this massive boost to the small size of the AID dataset, allowing our synthetic imagery to more effectively expand the limited training set.

\section{Conclusion}
\label{sec:conclusion}
We introduced \TerraDiTOmega, an end-to-end spatial control framework for GeoAI workflows that synthesizes high-fidelity satellite imagery directly from native geospatial primitives. Bridging the gap between expensive dense rasters and imprecise sparse prompts, our approach supports controllable layouts across varying annotation budgets through Geometry-Aware Local Attention, which injects explicit geometric cues tailored to each primitive format. \TerraDiTOmega{} outperforms existing baselines across all conditioning formats and serves as a versatile, unified framework for synthetic data augmentation, yielding consistent improvements in land-cover segmentation, object detection, road graph extraction, and scene classification. Currently, available geospatial datasets pair primitives with standard OSM tags rather than rich visual descriptions, so fine-grained attributes like color and texture are better controlled globally. As descriptive instance-level captions become available, extending our framework to primitive-level visual control is a natural next step. Furthermore, extending our data augmentation demonstrations to include other compatible downstream tasks, such as change detection, remains a compelling area for future research. Finally, we acknowledge that the ability to synthesize high-fidelity satellite imagery carries potential risks for downstream misuse, such as the generation of deceptive geographical data or the unintentional enhancement of harmful surveillance systems, underscoring the need for responsible deployment.

\section*{Acknowledgements}
This research used the TGI RAILs advanced compute and data resource, which is supported by the National Science Foundation (award OAC-2232860) and the Taylor Geospatial Institute. This work builds heavily from Sastry et al.~\cite{sastry2026terradit}.

\bibliographystyle{splncs04}
\bibliography{main}

\clearpage
\newpage
\beginsupplement
\title{\TerraDiTOmega: Unified Spatial Control for Satellite Image Synthesis with Any Geospatial Primitive \\ \normalsize -- Supplementary Material --}
\appendix
\section{Additional Implementation Details}

\subsection{Dataset Details}
\label{app:dataset}
\begin{figure}[ht]
    \centering

    \captionof{table}{\textbf{Dataset split statistics}. Distribution of images, instances, and tags across the training and testing splits for \TerraDiTOmega.}
    \label{tab:dataset_stats}
    \par\smallskip

    \resizebox{\linewidth}{!}{
    \begin{tabular}{l|c|c|c|c|c}
    Split & \# of Images & \# of Instances & \# of Empty Images & Unique Tags & Avg. Inst. \\
    \toprule
    Train              & 2M & 21.4M & 410K & 941 & 10.63 \\
    \midrule
    Test (Git-Rand)    & 15K & 162K & 3K & 453 & 10.84 \\
    Test (Git-Spatial) & 15K & 34K & 4K & 73 & 2.35 \\
    Test (Git-Dense)   & 3.5K & 128K & 0 & 421 & 37.41 \\
    \bottomrule
    \end{tabular}
    }

    \par\bigskip

    \includegraphics[width=0.55\linewidth]{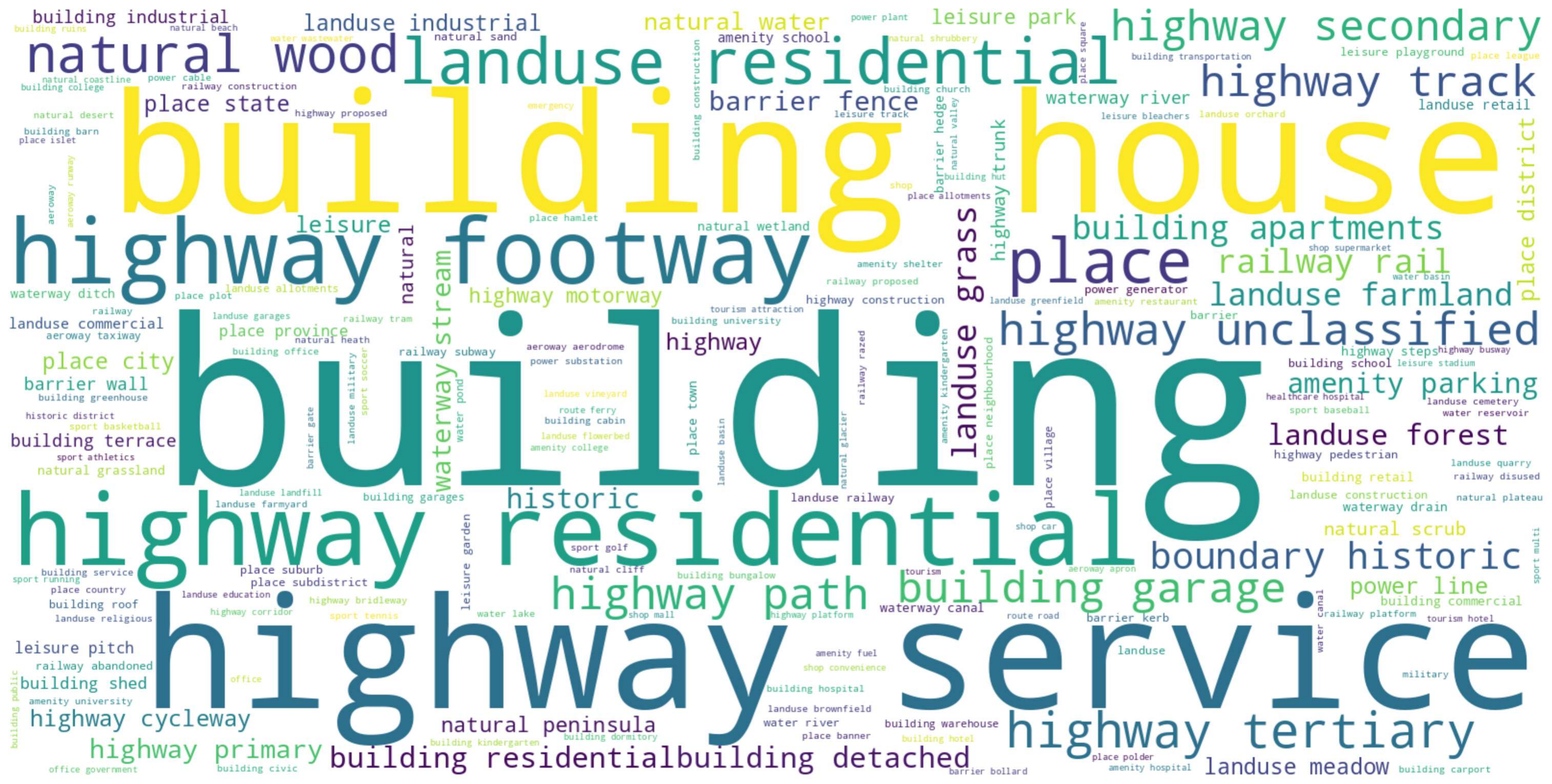}
    \par\smallskip
    \captionof{figure}{\textbf{Word cloud of semantic tags} used for instance captions. Larger words indicate more common tags.}
    \label{fig:tag_wordcloud}
\end{figure}

We extract our training dataset from Git-10M~\cite{liu2025text2earth} at zoom level 17 with $256 \times 256$ pixel resolution, utilizing semantic OSM tags directly as instance captions (Fig. \ref{fig:tag_wordcloud}). Compared to standard natural language or class-based vision datasets, our geospatial data is distinct: it contains a significantly larger vocabulary of unique tags, frequent spatial overlap among primitives, and a high concentration of very small instances in densely populated scenes.

To manage computational complexity, we cap each tile at a maximum of 64 instances and 64 vertices per geometry. While we do not find this cap limiting, such truncation could present challenges when scaling to lower zoom levels that encompass broader geographical areas. Therefore, scaling the Unified Primitive Encoder to efficiently handle higher capacities and more complex geometries remains a promising direction for future work.

\subsubsection{Testing Splits.}
\begin{figure}[ht]
    \centering
    \includegraphics[width=\linewidth]{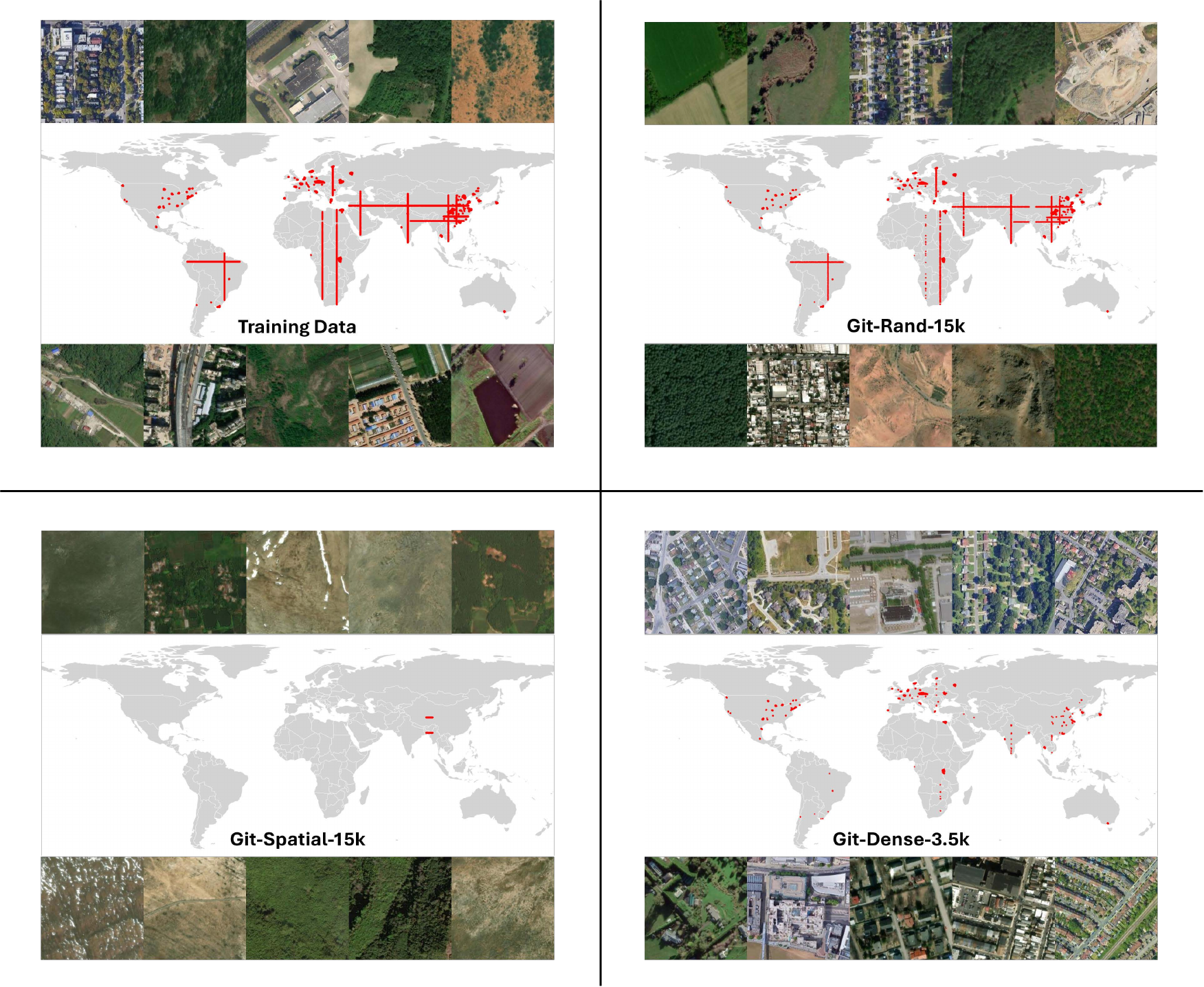}
    \caption{\textbf{Train and Test Splits.} We show the geolocation distribution and random image samples for each split.}
    \label{fig:dataset_splits}
\end{figure}

To comprehensively evaluate the model, we utilize three distinct testing splits (visualized in Fig. \ref{fig:dataset_splits}), with detailed statistics provided in Table \ref{tab:dataset_stats}. Following the evaluation protocol established by TerraDiT~\cite{sastry2026terradit}, we adopt the Git-Rand and Git-Spatial splits. Git-Rand captures the overall distribution of the dataset, maintaining natural sparsity by including thousands of empty images. Git-Spatial isolates a held-out geographic location to test zero-shot generalization; notably, this split is highly sparse (averaging 2.35 instances per tile), evaluating the model's robustness to minimal label conditioning and its generalist capabilities beyond strict layout control. To evaluate performance on more difficult scenes, we introduce a third split, Git-Dense. By filtering Git-Rand for tiles containing at least 15 instances, Git-Dense serves as a rigorous stress test for layout controllability in highly complex environments.

\subsection{Model Details}

\subsubsection*{Training Details.}
We utilize the TerraDiT-XL backbone~\cite{sastry2026terradit}, compressing images into a latent space via the frozen SDXL-VAE~\cite{podell2023sdxl}. Text encoding is performed using a frozen LongCLIP model~\cite{zhang2024long}, and Representation Alignment (REPA)~\cite{yu2024representation} is applied with a satellite-only DINOv3-ViT-L~\cite{simeoni2025dinov3} teacher model ($\lambda = 0.5$). The model is optimized using AdamW~\cite{loshchilov2017decoupled_adamw} ($\beta_1 = 0.9$, $\beta_2 = 0.999$, weight decay 0.0, $\epsilon = 1 \times 10^{-8}$) with a constant learning rate of $2 \times 10^{-5}$ and a global batch size of 256 distributed across 4 NVIDIA H100 GPUs. We employ a continuous-time diffusion framework with a linear noise schedule, velocity-prediction, and uniform loss weighting. During training, a 10\% independent dropout rate is applied to the geolocation embeddings, global captions, and all instance data. Furthermore, we apply dropout to specific primitive formats using a step-aware cosine annealing schedule: the dropout probability is held at $p = 0.5$ for the first 33\% of training steps, decays via a cosine curve, and remains clamped at $p = 0.1$ for the final 15\% of training. For ablations, we follow the same procedure, but with the B (base) model. The differences in the XL and B models are shown in Table~\ref{tab:sit_configs}.

\begin{table}[h]
    \centering
    \caption{\textbf{Architectural configurations}. Model specifications for the TerraDiT-XL and TerraDiT-B models.}
    \label{tab:sit_configs}
    \resizebox{0.6\linewidth}{!}{
    \begin{tabular}{l|c|c|c|c}
    Model & Depth & Hidden Size & Patch Size & Num Heads \\
    \toprule
    TerraDiT-XL & 28  & 1152 & 2 & 16 \\
    TerraDiT-B  & 12  & 768  & 2 & 12 \\
    \bottomrule
    \end{tabular}
    }
\end{table}

\subsubsection*{GALA Details.}
To ensure computational efficiency, the Spatial Geometry Field (SGF) operations are heavily vectorized in practice; we detail the implementation logic for bounding boxes, polylines, and polygons in Algorithms \ref{alg:sgf_bbox}, \ref{alg:sgf_polyline}, and \ref{alg:sgf_polygon}, respectively.

\renewcommand{\algorithmiccomment}[1]{\tabto{5cm}\textcolor{gray}{\# \textit{#1}}}

\begin{algorithm}[ht!]
\caption{Spatial Geometry Field (SGF) for Bounding Boxes}
\label{alg:sgf_bbox}
\textbf{Input}: Bounding box $B = (x_{\text{min}}, y_{\text{min}}, x_{\text{max}}, y_{\text{max}})$, Spatial grid $G$, Kernel function $\mathcal{K}$ \\
\textbf{Output}: Spatial Geometry Field $\mathbf{F}_{\text{bbox}}$
\begin{algorithmic}[1]
\STATE Initialize $\mathbf{F}_{\text{bbox}} \gets \mathbf{0}_{H \times W}$
\FOR{each pixel $(x, y) \in G$}
    \STATE $d_x \gets \max(x_{\text{min}} - x, 0, x - x_{\text{max}})$
    \STATE $d_y \gets \max(y_{\text{min}} - y, 0, y - y_{\text{max}})$
    \STATE $d \gets \sqrt{d_x^2 + d_y^2}$ \COMMENT{Euclidean distance to box boundary}
    \STATE $\mathbf{F}_{\text{bbox}}(x, y) \gets \mathcal{K}(d)$
\ENDFOR
\RETURN $\mathbf{F}_{\text{bbox}}$
\end{algorithmic}
\end{algorithm}

\begin{algorithm}[ht!]
\caption{Spatial Geometry Field (SGF) for Polylines}
\label{alg:sgf_polyline}
\textbf{Input}: Polyline segments $S = \{s_1, s_2, \dots, s_k\}$, Spatial grid $G$, Kernel function $\mathcal{K}$ \\
\textbf{Output}: Spatial Geometry Field $\mathbf{F}_{\text{line}}$
\begin{algorithmic}[1]
\STATE Initialize $\mathbf{F}_{\text{line}}$ with $\infty$ for all elements
\FOR{each pixel $(x, y) \in G$}
    \STATE $d_{\text{min}} \gets \infty$
    \FOR{each segment $s_i \in S$}
        \STATE $d_{\text{temp}} \gets \text{PointToSegmentDistance}((x,y), s_i)$
        \STATE $d_{\text{min}} \gets \min(d_{\text{min}}, d_{\text{temp}})$
    \ENDFOR
    \STATE $\mathbf{F}_{\text{line}}(x, y) \gets \mathcal{K}(d_{\text{min}})$
\ENDFOR
\RETURN $\mathbf{F}_{\text{line}}$
\end{algorithmic}
\end{algorithm}

\begin{algorithm}[ht!]
\caption{Spatial Geometry Field (SGF) for Polygons}
\label{alg:sgf_polygon}
\textbf{Input}: Polygon vertices $V = \{v_1, v_2, \dots, v_n\}$, Spatial grid $G$, Kernel function $\mathcal{K}$ \\
\textbf{Output}: Spatial Geometry Field $\mathbf{F}_{\text{poly}}$
\begin{algorithmic}[1]
\STATE Initialize $\mathbf{F}_{\text{poly}} \gets \mathbf{0}_{H \times W}$
\FOR{each pixel $(x, y) \in G$}
    \STATE $d_{\text{edge}} \gets \text{DistanceToNearestEdge}((x,y), V)$
    \STATE $m \gets \text{PointInPolygon}((x,y), V)$ \COMMENT{Returns $1$ if inside, $-1$ if outside}
    \STATE $d_{\text{signed}} \gets m \cdot d_{\text{edge}}$ \COMMENT{Compute signed distance}
    \STATE $\mathbf{F}_{\text{poly}}(x, y) \gets \mathcal{K}(d_{\text{signed}})$
\ENDFOR
\RETURN $\mathbf{F}_{\text{poly}}$
\end{algorithmic}
\end{algorithm}

\subsubsection*{Inference Details.}
To maintain consistency with TerraDiT-$\alpha$ and TerraDiT-$\Sigma$, we adopt a similar inference procedure for \TerraDiTOmega. Specifically, we use a linear velocity prediction over 100 flow steps with a batch size of 32 and a CFG scale of 0.0. To isolate specific spatial representations during inference, we apply targeted masking: for point-only generation, we mask out bounding box, polygon, and polyline primitives, and for bounding-box-only generation, we mask out polygons and polylines.

\subsection{Metrics}

\subsubsection*{Fr\'echet Inception Distance.} To measure the visual fidelity and diversity of generated images, we employ Fr\'echet Inception Distance (FID)~\cite{heusel2017gans}. FID measures the similarity between generated and real images by computing the Fr\'echet distance between feature vectors extracted from both image sets using a pretrained Inception-v3 model. For our evaluation, we use \texttt{TorchMetrics}~\cite{detlefsen2022torchmetrics} to ensure consistency.

\subsubsection*{Spatial FID.} To measure the spatial coherence and structural fidelity of the generated images, we use Spatial Fréchet Inception Distance (sFID)~\cite{nash2021generating}. Unlike standard FID, which relies on average-pooled features that discard spatial information, sFID better captures spatial relationships by computing the Fr\'echet distance over intermediate spatial feature maps. In our PyTorch implementation, we extract features from the \texttt{Mixed\_6e} layer of an ImageNet-pretrained Inception-v3 model, slice the first seven channels, and flatten the spatial dimensions to compute the distribution distance between the real and generated images.

\subsubsection*{Learned Perceptual Image Patch Similarity.} We employ Learned Perceptual Image Patch Similarity (LPIPS)~\cite{zhang2018perceptual} to measure the perceptual distance between a real image and its generated counterpart. We employ the SqueezeNet variant from the \texttt{TorchMetrics}~\cite{detlefsen2022torchmetrics} implementation, which extracts features across layers and computes a weighted $L_2$ difference between them.

\subsubsection*{Structural Similarity Index Measure.} To measure the structural adherence of generated images, we use Structural Similarity Index Measure (SSIM)~\cite{wang2004image}, which computes the difference in luminance, contrast, and structure between corresponding generated and real images. We employ the \texttt{TorchMetrics}~\cite{detlefsen2022torchmetrics} implementation, which uses a sliding Gaussian window to compute local similarity maps.

\subsubsection*{CLIPScore.} To measure the semantic alignment between generated images and the corresponding text prompt, we use CLIPScore~\cite{hessel-etal-2021-clipscore}, which projects the text and image into a shared embedding space and computes their cosine similarity. We use a ViT-L/14 in our implementation.

\subsubsection*{Classification Accuracy Score.} To measure how well generated images follow spatial layouts, we compute the Classification Accuracy Score (CAS)~\cite{ravuri2019classification}. Following LayoutDiffusion~\cite{zheng2023layoutdiffusion}, we train a ResNet-110~\cite{he2016deep} from scratch on generated bounding box crops and evaluate on real crops. To preserve the aspect ratios of varying geometries, we zero-pad boxes before interpolating to $32 \times 32$ resolution, discarding instances smaller than 8 pixels. We also simplify the tag distribution by collapsing OSM tags into 13 parent classes (e.g., `building' for `building apartment'). For training on the Git-Rand-15k set, we use random cropping and horizontal flipping augmentations, a batch size of 128, and optimize via SGD for 160 epochs (initial learning rate 0.1, decayed by 0.1 at epochs 80 and 120). The best-performing checkpoint is selected using a 10\% validation split.
\section{Additional Qualitative Results}
\begin{figure}[tb]
    \centering
    \includegraphics[width=0.6\linewidth]{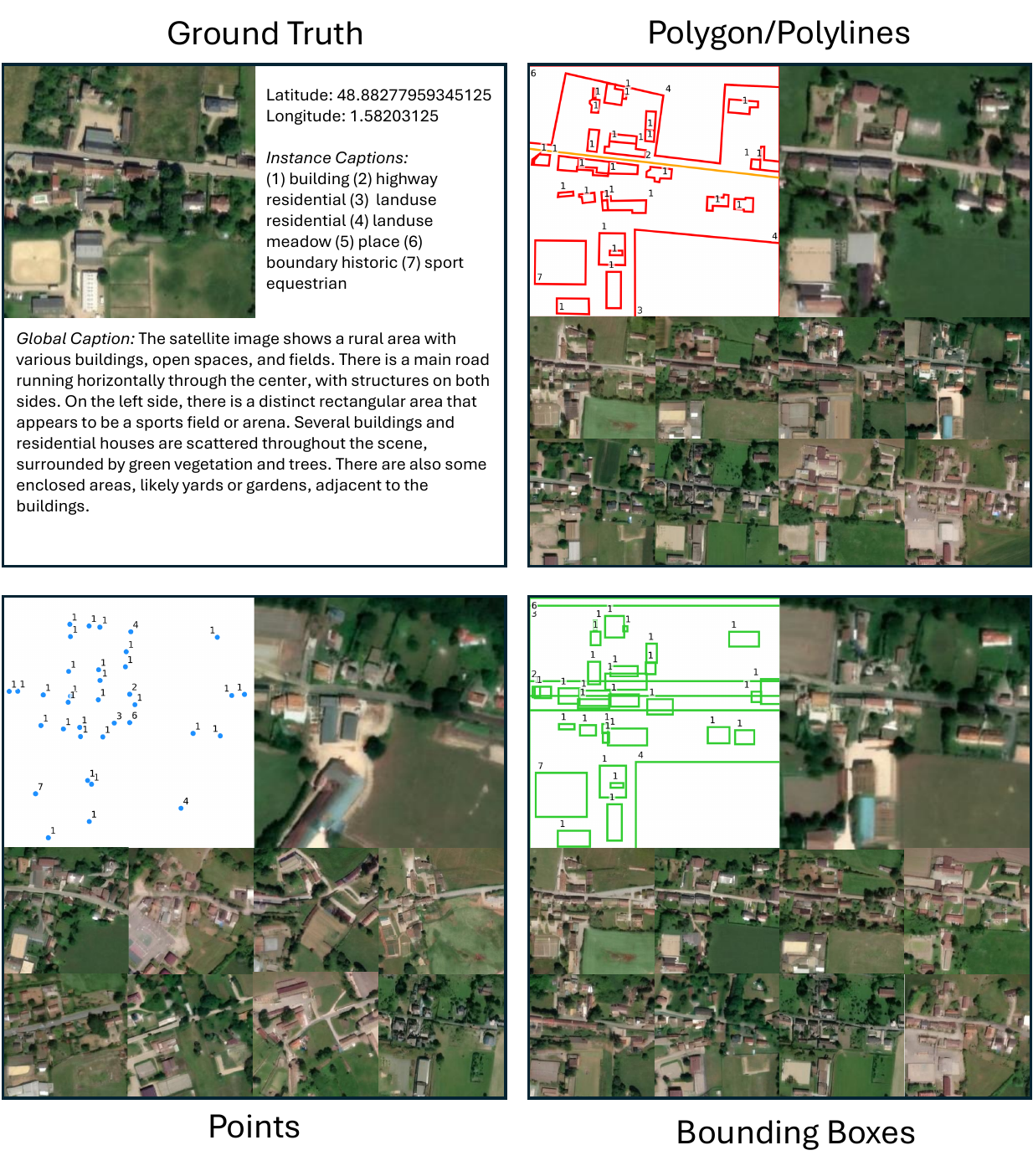}
    \caption{\TerraDiTOmega{}-XL Uncurated Samples. Rural area in Tilly, France.}
    \label{fig:samples_5}

    \par\medskip

    \includegraphics[width=0.6\linewidth]{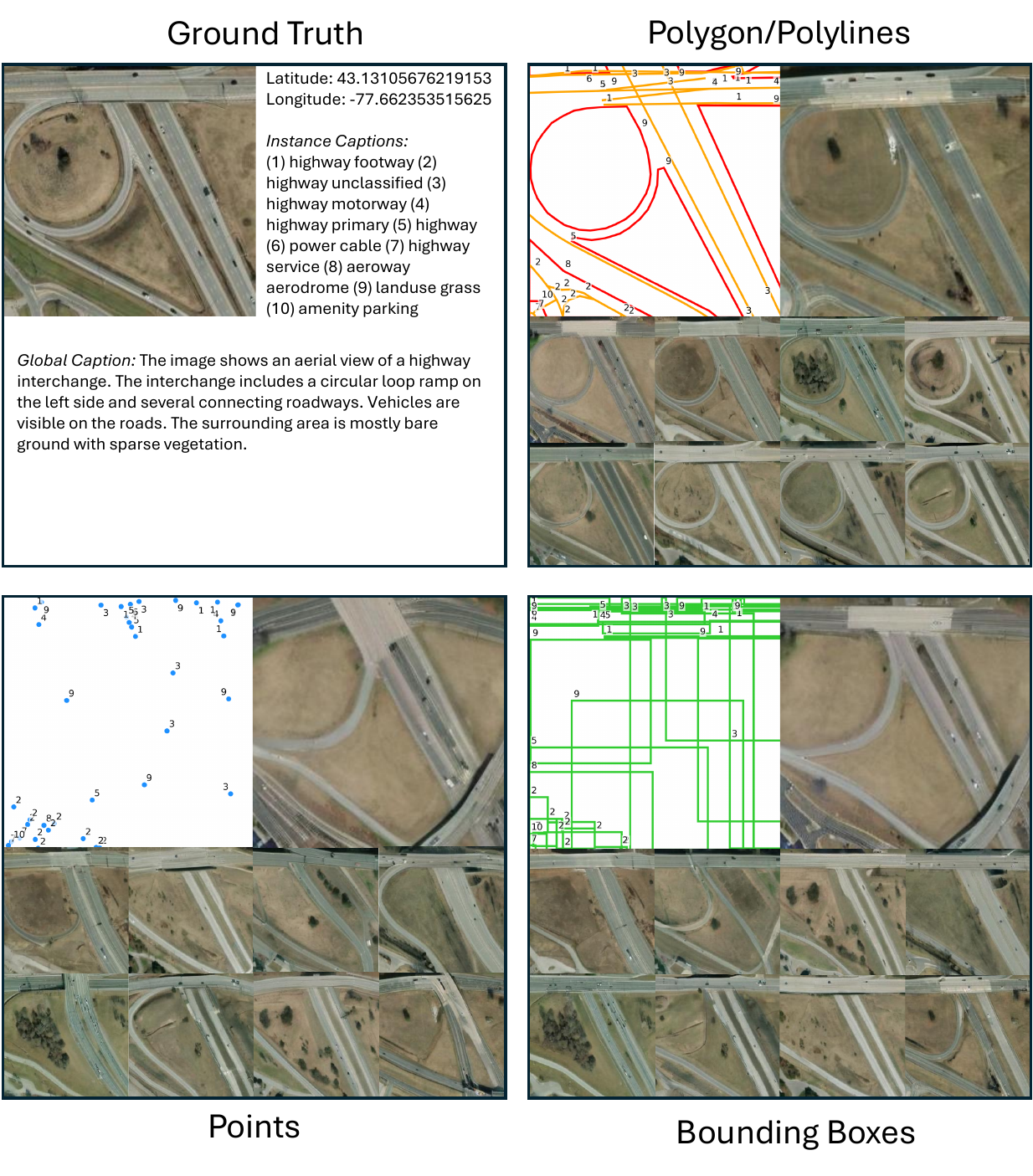}
    \caption{\TerraDiTOmega{}-XL Uncurated Samples. Highway interchange in New York, USA.}
    \label{fig:samples_6}
\end{figure}

\begin{figure}[tb]
    \centering
    \includegraphics[width=\linewidth]{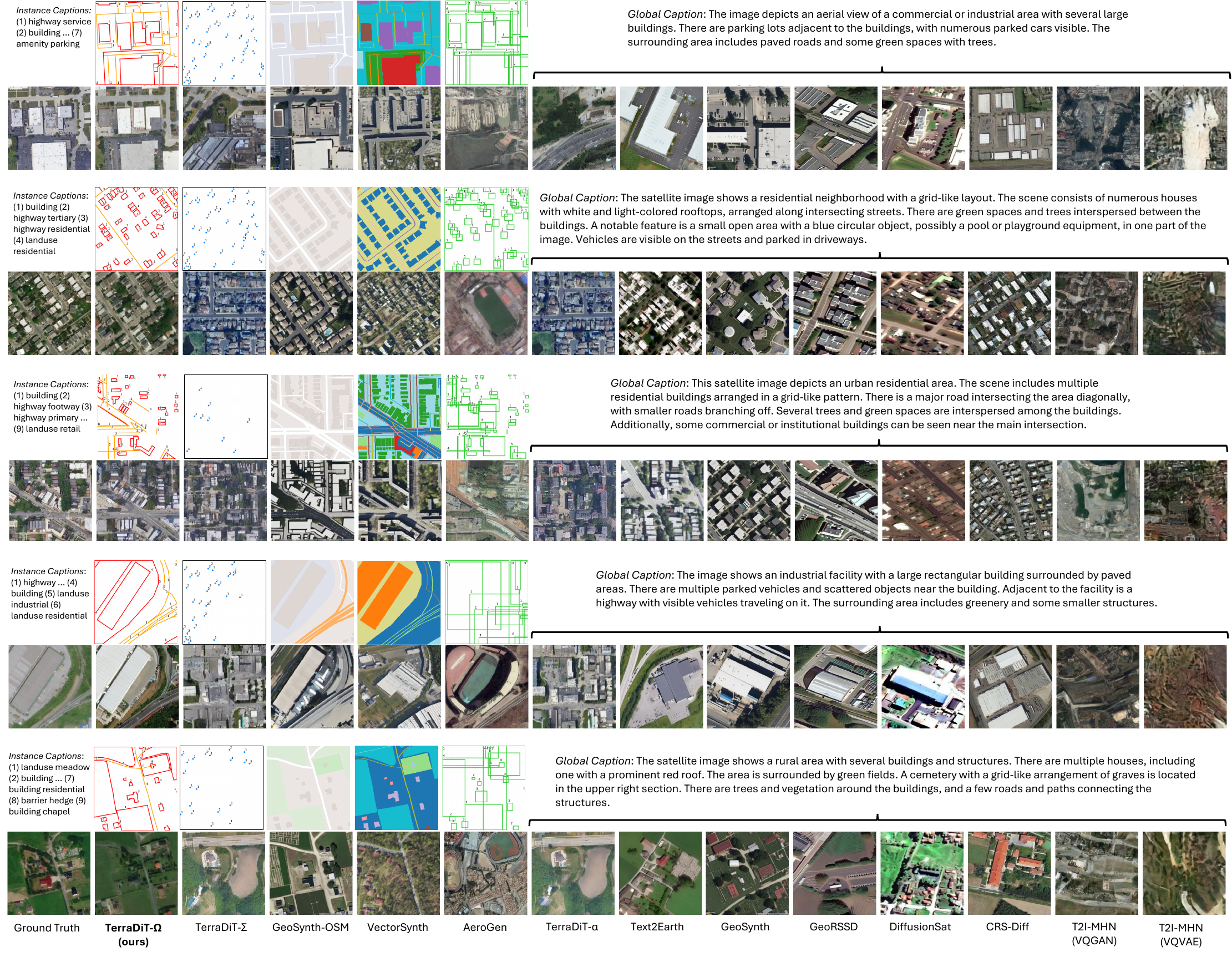}
    \caption{Qualitative Comparisons with Geospatial Models.}
    \label{fig:geospatial_comparisons}
\end{figure}

\begin{figure}[tb]
    \centering
    \includegraphics[width=\linewidth]{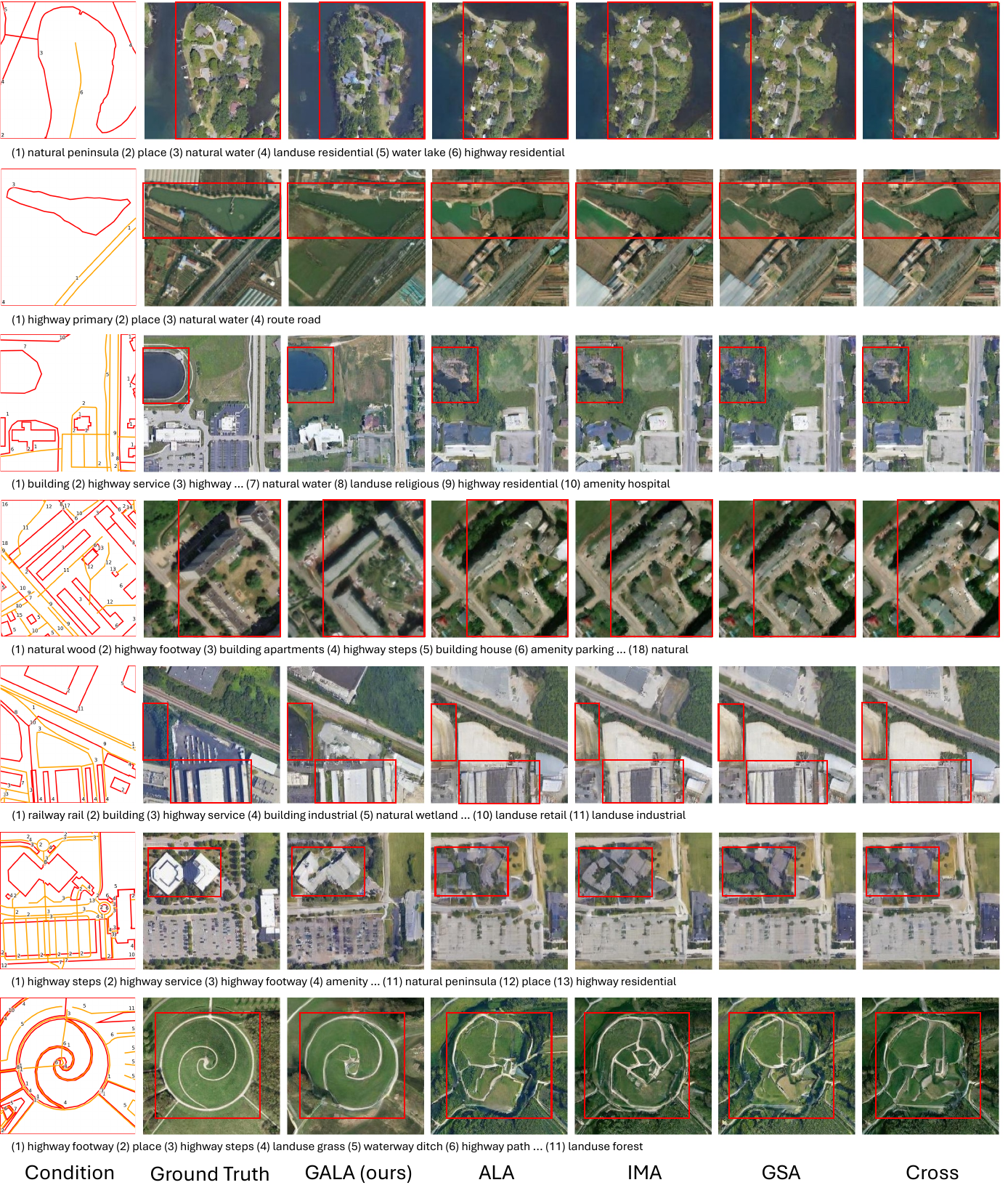}
    \caption{More Comparisons of Attention Methods (B/2 Models). GALA is our Geometry-Aware Local Attention, ALA is Adaptive Local Attention~\cite{sastry2026terradit}, IMA is Instance Masked Attention~\cite{wang2024instancediffusion}, GSA is Gated Self-Attention~\cite{li2023gligen}, and Cross is vanilla cross-attention.}
    \label{fig:attention_comparison_app}
\end{figure}

We provide uncurated samples with varying textual, geolocation, and spatial conditions in Fig~\ref{fig:samples_1} to \ref{fig:samples_6}. We also show more qualitative comparisons to geospatial models in Fig.~\ref{fig:geospatial_comparisons}. For robust comparisons on difficult geometries, we compare our ablated base models (all trained from scratch for 400k steps on the same backbone and data) in Fig.~\ref{fig:attention_comparison_app}.

\section{Additional Quantitative Results}

\subsection{Further Model Comparisons.}

We additionally evaluate structural fidelity and text-alignment using SSIM and CLIPScore (Table~\ref{tab:appendix_quant}). While our model achieves competitive metrics across all splits, we hypothesize that CLIP aligns poorly with both overhead imagery and the dense captions of Git-10M. Notably, general models still score well on this metric despite failing to generate realistic satellite imagery. When evaluating structural generation (SSIM), the Git-Rand-15k and Git-Spatial-15k splits contain numerous sparse or unannotated tiles, such as barren terrain or forests. Consequently, the Git-Dense-3.5k split, which features highly complex scenes, serves as a much more rigorous benchmark for layout-conditioned generation. On this dense split, \TerraDiTOmega-XL significantly outperforms all baselines. Furthermore, we observe a distinct scaling behavior in structural fidelity: SSIM consistently improves as annotation complexity increases from points, to bounding boxes, and ultimately to polygons and polylines.

\begin{table*}[!t]
\centering
\caption{\textbf{Model Details and Zero-shot evaluation on Git-Rand-15k, Git-Spatial-15k, and Git-Dense-3.5k.}
$\uparrow$ indicates higher is better.
$T$ is text input, $O$ is OSM raster input, $L$ is geolocation, $P$ is points, $B$ is bounding box, and $\Omega$ is \{polygon, polyline, bounding box, and point conditioning\}.
\scriptsize{
\textit{*Text2Earth was trained on the entire Git-10M dataset (including our test sets), which may inflate reported performance.}}
}
\label{tab:appendix_quant}
\setlength{\aboverulesep}{0pt}
\setlength{\belowrulesep}{0pt}
\resizebox{\linewidth}{!}{
\begin{tabular}{l|c|c|c|c|cc|cc|cc}
\multirow{2}{*}{Model} & \multirow{2}{*}{Condition} &
\multirow{2}{*}{Training Set} & \multirow{2}{*}{Images} & \multirow{2}{*}{Params} &
\multicolumn{2}{c|}{\textbf{Git-Rand-15k}} &
\multicolumn{2}{c|}{\textbf{Git-Spatial-15k}} &
\multicolumn{2}{c}{\textbf{Git-Dense-3.5k}} \\
\cline{6-11}
& & & & & SSIM $\uparrow$ & CLIP $\uparrow$ & SSIM $\uparrow$ & CLIP $\uparrow$ & SSIM $\uparrow$ & CLIP $\uparrow$ \\
\toprule
\rowcolor{gray!25}
\multicolumn{11}{l}{\textit{\textbf{General} Models}} \\
SDXL~\cite{podell2023sdxl} & $T$ & Internal & N/A & 2.6B &
0.1817 & 0.2423 & 0.2597 & 0.2389 & 0.1004 & 0.2490 \\
SD 3~\cite{esser2024scaling} & $T$ & Internal & N/A & 2B &
0.1598 & 0.2567 & 0.2000 & 0.2691 & 0.1012 & 0.2639 \\
PixArt-$\alpha$-XL~\cite{chen2023pixart} & $T$ & ImageNet/SAM/Internal & 25M & 611M &
0.1175 & 0.2509 & 0.1738 & 0.2525 & 0.0669 & 0.2657 \\
PixArt-$\Sigma$-XL~\cite{chen2024pixart} & $T$ & Internal & 33M & 611M &
0.1574 & 0.2535 & 0.2271 & 0.2683 & 0.0912 & 0.2631 \\
InstanceDiffusion~\cite{wang2024instancediffusion} & $T + B$ & Internal & 5M & 1.23B &
0.0957 & 0.2063 & 0.1203 & 0.2119 & 0.0772 & 0.2249 \\
GLIGEN~\cite{li2023gligen} & $T + B$ & LVIS/GoldG/O365/SBU/CC3M & 7M & 1.07B &
0.1658 & 0.2703 & 0.2109 & 0.2751 & 0.0949 & 0.2649 \\
\midrule
\rowcolor{gray!25}
\multicolumn{11}{l}{\textit{\textbf{Geospatial} Models}} \\
MHN-VQGAN~\cite{xu2023txt2img} & $T$ & RSICD & 9K & 116M &
0.1482 & 0.2170 & 0.1747 & 0.1865 & 0.1032 & 0.2478 \\
MHN-VQVAE~\cite{xu2023txt2img} & $T$ & RSICD & 9K & 29M &
0.1729 & 0.2451 & 0.2074 & 0.2450 & 0.1144 & 0.2173 \\
GeoRSSD~\cite{zhang2024rs5m} & $T$ & RS5M & 5M & 866M &
0.1780 & 0.2692 & 0.2443 & 0.2549 & 0.1049 & 0.2798 \\
CRS-Diff~\cite{tang2024crs} & $T$ & RSICD/fMoW/MillionAID & 1.1M & 860M &
0.1855 & 0.2798 & 0.2306 & 0.2715 & 0.1138 & \underline{0.2854} \\
DiffusionSat~\cite{khanna2023diffusionsat} & $T+L$ & fMoW/Satlas/SpaceNet & 2M & 880M &
0.1615 & 0.2597 & 0.2166 & 0.2430 & 0.0983 & 0.2803 \\
GeoSynth~\cite{sastry2024geosynth} & $T$ & Internal & 45K & 866M &
0.1973 & 0.2478 & 0.2458 & 0.2521 & 0.1178 & 0.2649 \\
GeoSynth-OSM~\cite{sastry2024geosynth} & $T+O$ & Internal & 45K & 1.23B &
0.1979 & 0.2629 & 0.2135 & 0.2011 & 0.1450 & 0.2705 \\
VectorSynth~\cite{cher2026vectorsynth} & $T+O$ & OSM-Polygon & 20K & 1.23B &
0.1711 & 0.2763 & 0.2110 & 0.2646 & 0.1111 & 0.2842 \\
AeroGen~\cite{tang2025aero} & $T + B$ & DIOR/-R & 45K & 907M &
0.2049 & 0.2153 & 0.2535 & 0.2134 & 0.1526 & 0.2095 \\
Text2Earth$^*$~\cite{liu2025text2earth} & $T$ & Git-10M & 10M & 867M &
0.2371 & \underline{0.2871} & 0.3219 & 0.2743 & 0.1349 & \textbf{0.2914} \\
TerraDiT-$\alpha$-XL~\cite{sastry2026terradit} & $T$ & Git-10M & 2M & 836M &
0.2599 & 0.2787 & 0.3316 & \underline{0.2849} & 0.1275 & 0.2661 \\
TerraDiT-$\Sigma$-XL~\cite{sastry2026terradit} & $T+P+L$ & Git-10M/OSM & 2M & 1.12B &
\textbf{0.2740} & \textbf{0.2877} & \textbf{0.3497} & \textbf{0.2902} & 0.1329 & 0.2671 \\
\midrule
& $T+P+L$& & & &
0.2574 & 0.2804 & 0.3424 & 0.2761 & 0.1503 & 0.2787 \\
\textbf{TerraDiT-\TextOmega-XL (ours)} & $T+B+L$ & Git-10M/OSM & 2M & 1.18B &
0.2623 & 0.2813 & \underline{0.3427} & 0.2763 & \underline{0.1656} & 0.2813 \\
& $T+\Omega+L$ & & & &
\underline{0.2665} & 0.2812 & 0.3421 & 0.2761 & \textbf{0.1757} & 0.2821 \\
\bottomrule
\end{tabular}
}
\end{table*}

\begin{figure}[H]
    \centering
    \includegraphics[width=0.6\linewidth]{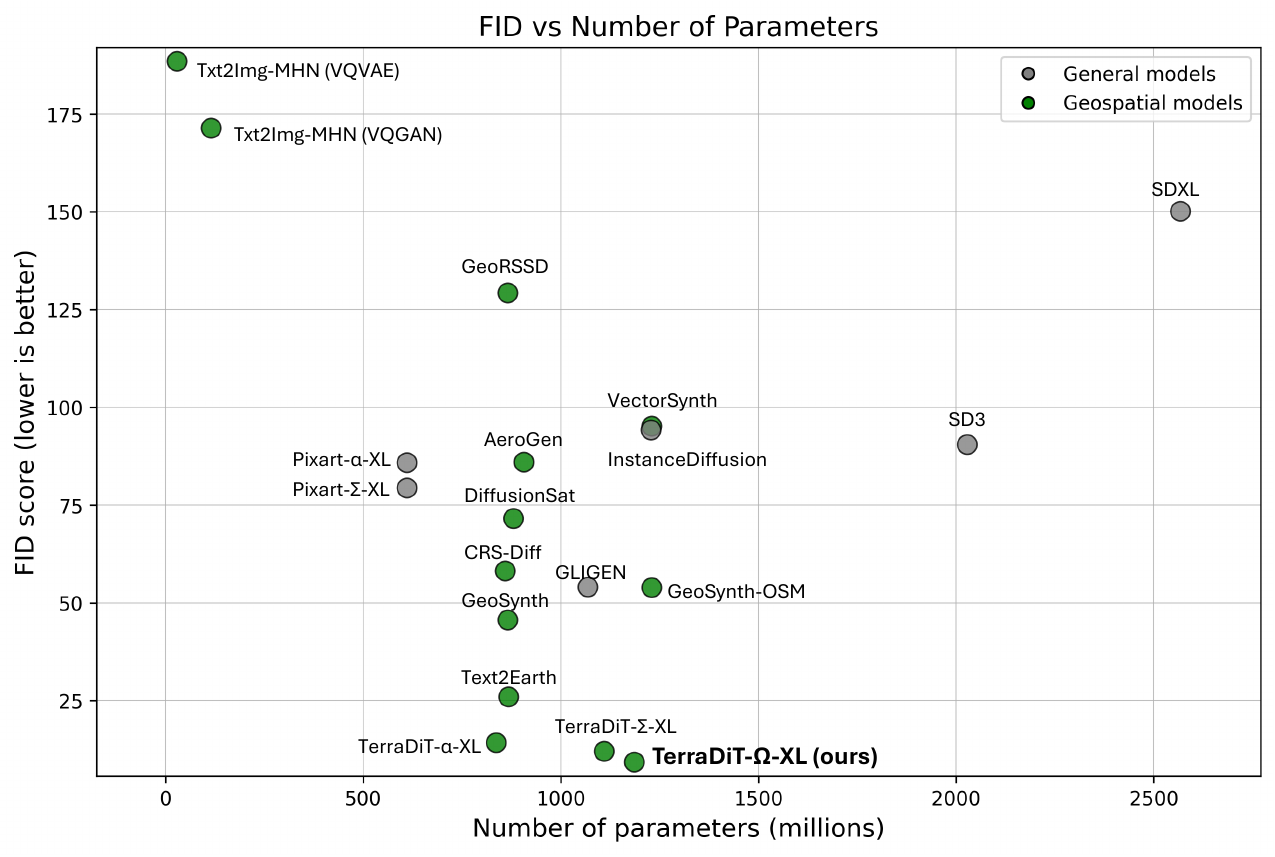}
    \caption{Comparison with baseline approaches on zero-shot evaluation on Git-Rand-15k. Our approach exhibits best FID with comparable parameter count.}
    \label{fig:fid_vs_params}
\end{figure}

Regarding model architecture and efficiency, our parameter count remains comparable to baselines. As seen in Table~\ref{tab:appendix_quant} and Fig.~\ref{fig:fid_vs_params}, \TerraDiTOmega-XL (1.18B) operates with fewer parameters than ControlNet-based geospatial models like GeoSynth-OSM~\cite{sastry2024geosynth} and VectorSynth~\cite{cher2026vectorsynth}, as well as general layout models like InstanceDiffusion~\cite{wang2024instancediffusion}. While general vision models are trained on vastly larger datasets, their undisclosed distributions often result in poor generalization to the overhead satellite domain. Conversely, most layout-based geospatial methods are bottlenecked by the scarcity of dense annotations and train on significantly smaller datasets. For direct in-domain comparisons on Git-10M, primary baselines are Text2Earth~\cite{liu2025text2earth}, TerraDiT-$\alpha$-XL~\cite{sastry2026terradit}, and TerraDiT-$\Sigma$-XL~\cite{sastry2026terradit}. However, it is important to note that Text2Earth was trained on the entirety of Git-10M, including our held-out test sets, which may artificially inflate its reported performance metrics.

\subsection{Primitive Data Efficiency Analysis}
In standard computer vision workflows, instance masks are typically precomputed into rasters for compatibility with traditional dataloaders. However, Remote Sensing (RS) scenes are vastly denser. Precomputing millions of individual mask files introduces practical workflow inefficiencies, such as increased storage overhead and additional data-conversion. To evaluate scalable alternatives, we benchmarked three dataloading strategies on 100 randomly sampled tiles from the Git-Dense-3.5k dataset: (1) direct primitive loading, which utilizes vectorized NumPy slicing to extract instance geometries directly from compressed metadata arrays; (2) precomputed raster loading, measuring the disk I/O latency of sequentially reading cached masks; (3) on-the-fly rasterization, evaluating the computational cost of dynamically rendering geometries into empty mask tensors.

\begin{table}[h]
    \centering
    \caption{\textbf{Dataloading Latency Comparison}.}
    \label{tab:loading_latency}
    \resizebox{0.4\linewidth}{!}{
    \begin{tabular}{l|c}
    Data & Latency (ms) \\
    \toprule
    Direct (primitives) & 1007.0$\pm$14.7  \\
    Precomputed Rasters & 1017.0$\pm$96.0  \\
    Rasterization & 5575.8$\pm$60.5 \\
    \bottomrule
    \end{tabular}
    }
\end{table}

As shown in Table~\ref{tab:loading_latency}, while precomputed rasters (1017.0 ms) and direct primitives (1007.0 ms) yield similar latencies, the precomputed route imposes a data-management cost (storage and precomputation time). Alternatively, while on-the-fly rasterization circumvents storage friction, it introduces a severe latency bottleneck, taking 5575.8 ms even at a modest 256 $\times$ 256 resolution. Consequently, directly loading primitives emerges as the most seamless approach for dense GeoAI pipelines. This advantage becomes critical at higher resolutions; storage requirements for precomputed masks and the computational cost of on-the-fly rasterization both grow quadratically. In contrast, the complexity of loading primitives remains constant regardless of target resolution, making it a highly scalable choice.
\subsection{Data Augmentation}
In the following subsection, we provide further implementation details regarding our synthetic data augmentation experiments. We also include alternative baselines for object detection and land-cover segmentation.

For all tasks, we fine-tune our \TerraDiTOmega-XL model on the training dataset for 15K steps with an effective batch size of 128 across two NVIDIA A100 GPUs. To construct spatial conditions for synthetic training sets, we extract layouts from the real dataset and randomly apply seven transformations (flips, rotations, and transposes). For OpenEarthMap, we also include geolocation conditioning.

\subsubsection*{DIOR.}
The DIOR dataset~\cite{Li_2020} contains annotations for 20 object classes at varying spatial resolutions, which we partition into 18K, 2K, and 3.5K images for training, validation, and testing, respectively, with all imagery resized to 256 $\times$ 256 pixels.

For the downstream object detection task, we train an RT-DETR model~\cite{zhao2024detrs} from scratch for 130 epochs with a batch size of 32 across 2 NVIDIA A100 GPUs. The architecture utilizes a randomly initialized ResNet-50 backbone~\cite{he2016deep}, and is trained using the AdamW~\cite{loshchilov2017decoupled_adamw} optimizer with a learning rate of $1 \times 10^{-4}$ and a weight decay of $1 \times 10^{-4}$, following the original implementation. We use the validation set to determine our best model to ensure convergence.
\begin{table}
    \centering
    \caption{\textbf{Synthetic data augmentation for object detection.} Synthetic samples undergo no postprocessing. Our approach still shows improvements while the AeroGen~\cite{tang2025aero} drastically hurts performance with increased synthetic training. We use RT-DETR~\cite{zhao2024detrs} for object detection on the DIOR dataset.}
    \label{tab:dior_synth}
    \resizebox{0.75\linewidth}{!}{
    \begin{tabular}{c|cc|cc}
    \multirow{2}{*}{\makecell{Synth \\ Data}} & \multicolumn{2}{c|}{\TerraDiTOmega-XL (ours)} & \multicolumn{2}{c}{AeroGen~\cite{tang2025aero}} \\
    \cline{2-5}
     & mAP@50-95 $\uparrow$ & mAP@50 $\uparrow$ & mAP@50-95 $\uparrow$ & mAP@50 $\uparrow$ \\
    \toprule
    $\times0$& 55.15$\pm$0.15 & 77.09$\pm$0.31 & 55.15$\pm$0.15 & 77.09$\pm$0.31 \\
    $\times1$& \textbf{56.14$\pm$0.18} & 78.69$\pm$0.15 & 52.15$\pm$0.39 & 74.00$\pm$0.54 \\
    $\times2$& 55.90$\pm$0.17 & \textbf{78.87$\pm$0.28} & 47.20$\pm$0.06 & 68.29$\pm$0.17 \\
    \bottomrule
    \end{tabular}
    }
\end{table}

\noindent We establish a baseline for synthetic data augmentation on the DIOR dataset to evaluate raw generation quality. Our approach yields improvements in downstream object detection directly from the generated imagery, circumventing the need for additional post-processing. In contrast, methods such as AeroGen~\cite{tang2025aero} employ a multi-stage filtering pipeline—including synthetic layout generation, semantic filtering via CLIP, and layout consistency validation using bounding box classifiers. While this filtering process ensures quality control, it discards a substantial portion of the generated samples and depends on undisclosed thresholds, which increases the computational overhead for large-scale dataset synthesis. To ensure a standardized comparison, we generate images from AeroGen's public checkpoint using our exact augmented layouts without applying post-processing. As shown in Table \ref{tab:dior_synth}, without its specific filtering pipeline, the raw synthetic imagery from AeroGen decreases downstream performance. Conversely, our model consistently improves RT-DETR metrics out-of-the-box, demonstrating the strong baseline alignment of our generated data.

\subsubsection*{OpenEarthMap.}
The OpenEarthMap dataset~\cite{xia2023openearthmap} contains 5K high-resolution $1024 \times 1024$ images spanning 97 regions and 44 countries, annotated with 8 land-cover categories. We preprocess these images by extracting non-overlapping $256 \times 256$ patches. We use the official validation set for testing and randomly split the training data 90/10 for model selection. To adapt our model for land-cover segmentation on the OpenEarthMap dataset, we convert the dense segmentation maps into polygonal formats to align with our geometric primitive conditioning.

To evaluate the utility of the synthesized imagery, we train a U-Net~\cite{ronneberger2015u} segmentation model with a ResNet-34 backbone~\cite{he2016deep} from scratch. The model is trained for 100 epochs using the AdamW~\cite{loshchilov2017decoupled_adamw} optimizer with a learning rate of $1 \times 10^{-4}$ and a weight decay of $1 \times 10^{-6}$, utilizing the validation set to select the best-performing weights.

While prior works such as SatSynth~\cite{toker2024satsynth} conduct similar synthetic data evaluations, the lack of open-source code precludes direct replication. Therefore, to establish a robust comparative baseline, we fine-tune GeoSynth~\cite{sastry2024geosynth} using ControlNet-style~\cite{zhang2023adding} conditioning directly on the rasterized OpenEarthMap segmentation maps. We maintain the exact same training configuration and dataset as our model—training for 15K steps across two A100 GPUs with an effective batch size of 128, utilizing gradient accumulation to manage memory constraints. The inference procedure is identically configured to ensure a fair comparison.

\begin{table}
    \centering
    \caption{\textbf{Synthetic data augmentation for land-cover segmentation.} Synthetic samples undergo no postprocessing. Our approach shows slight improvements over GeoSynth~\cite{sastry2024geosynth} finetuned via ControlNet-style~\cite{zhang2023adding} conditioning. We use U-Net~\cite{ronneberger2015u} for segmentation on the OpenEarthMap dataset.}
    \label{tab:oem_synth}
    \resizebox{0.75\linewidth}{!}{
    \begin{tabular}{c|cc|cc}
    \multirow{2}{*}{\makecell{Synth \\ Data}} & \multicolumn{2}{c|}{\TerraDiTOmega-XL (ours)} & \multicolumn{2}{c}{GeoSynth~\cite{sastry2024geosynth} (tuned)} \\
    \cline{2-5}
     & mIoU $\uparrow$ & F1 $\uparrow$ & mIoU $\uparrow$ & F1 \\
    \toprule
    $\times0$& 55.75$\pm$0.26 & 70.70$\pm$0.24 & 55.75$\pm$0.26 & 70.70$\pm$0.24 \\
    $\times1$& 57.05$\pm$0.14 & 71.77$\pm$0.10 & 56.84$\pm$0.24 & 71.59$\pm$0.19 \\
    $\times2$& \textbf{57.46$\pm$0.12} & \textbf{72.01$\pm$0.12} & 57.30$\pm$0.21 & 72.00$\pm$0.20 \\
    \bottomrule
    \end{tabular}
    }
\end{table}

As shown in Table~\ref{tab:oem_synth}, while \TerraDiTOmega-XL yields minor quantitative gains over the fine-tuned GeoSynth baseline, the downstream segmentation performance between the two is highly comparable. However, the critical distinction lies in architectural flexibility. ControlNet-style conditioning is inherently tied to dense raster formats. In contrast, by natively processing geometric primitives, our model easily scales to diverse annotation types—such as polyline-based road graph extraction and bounding box-based object detection—completely bypassing the rigid inconvenience and computational overhead of the expensive rasterization steps required by traditional layout-to-image methods.

\subsubsection*{City-Scale.}
The City-Scale dataset~\cite{he2020sat2graph} comprises 180 high-resolution 2048 $\times$ 2048 satellite images spanning 20 U.S. cities, which we partition into 142, 9, and 29 images for training, validation, and testing, respectively (following SAM-Road~\cite{hetang2024segment}). We preprocess these high-resolution images by extracting non-overlapping 256 $\times$ 256 patches. To adapt our model for road graph extraction on the City-Scale dataset, we first segment the continuous road networks into separable polylines.

For the downstream road extraction task, we fine-tune the SAM-Road architecture~\cite{hetang2024segment} for 35 epochs with a batch size of 32 on a single NVIDIA A100 GPU. While the SAM backbone (ViT-B) is initialized with pre-trained weights, the decoders are trained from scratch. We use the Adam optimizer~\cite{kingma2015adam} with a base learning rate of $1 \times 10^{-3}$, following the original implementation. We use the validation set to determine our best model to ensure convergence, and we apply the thresholding values from the original SAM-Road implementation during inference.

\subsubsection*{AID.}
The AID dataset~\cite{xia2017aid} comprises 10K globally diverse satellite images distributed across 30 distinct scene categories, which we partition into a stratified split of 8K, 1K, and 1K images for training, validation, and testing, respectively. To construct the synthetic downstream training set, we generate class-balanced synthetic imagery at a resolution of $256 \times 256$.

For the downstream classification task, we train a ResNet-34 model~\cite{he2016deep} from scratch for 90 epochs using a batch size of 256. The model is optimized using SGD, a weight decay of $1 \times 10^{-4}$, and a base learning rate of 0.1. We employ a multi-step learning rate scheduler that applies a decay factor of 0.1 at training milestones of 30 and 60 epochs. Consistent with our other downstream evaluations, we utilize the validation set to select the best-performing model weights.

\renewcommand{\topfraction}{0.95}
\renewcommand{\bottomfraction}{0.95}
\renewcommand{\textfraction}{0.05}
\renewcommand{\floatpagefraction}{0.9}

\clearpage
\section{Visualizations}

\begin{center}
    \nopagebreak
    \includegraphics[width=0.6\linewidth]{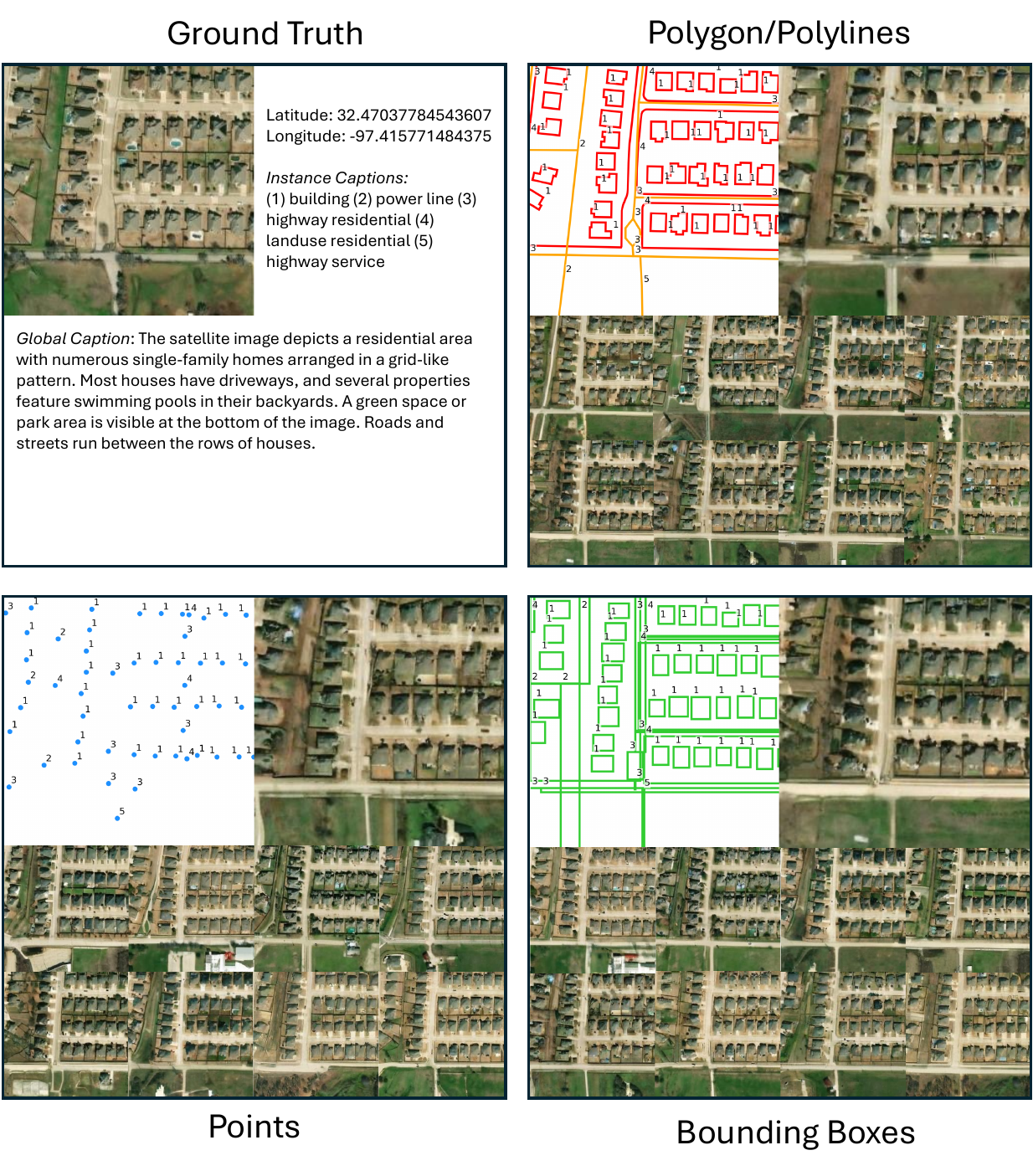}
    \captionof{figure}{\TerraDiTOmega{}-XL Uncurated Samples. Residential area in Texas, USA.}
    \label{fig:samples_1}

    \par\medskip

    \includegraphics[width=0.6\linewidth]{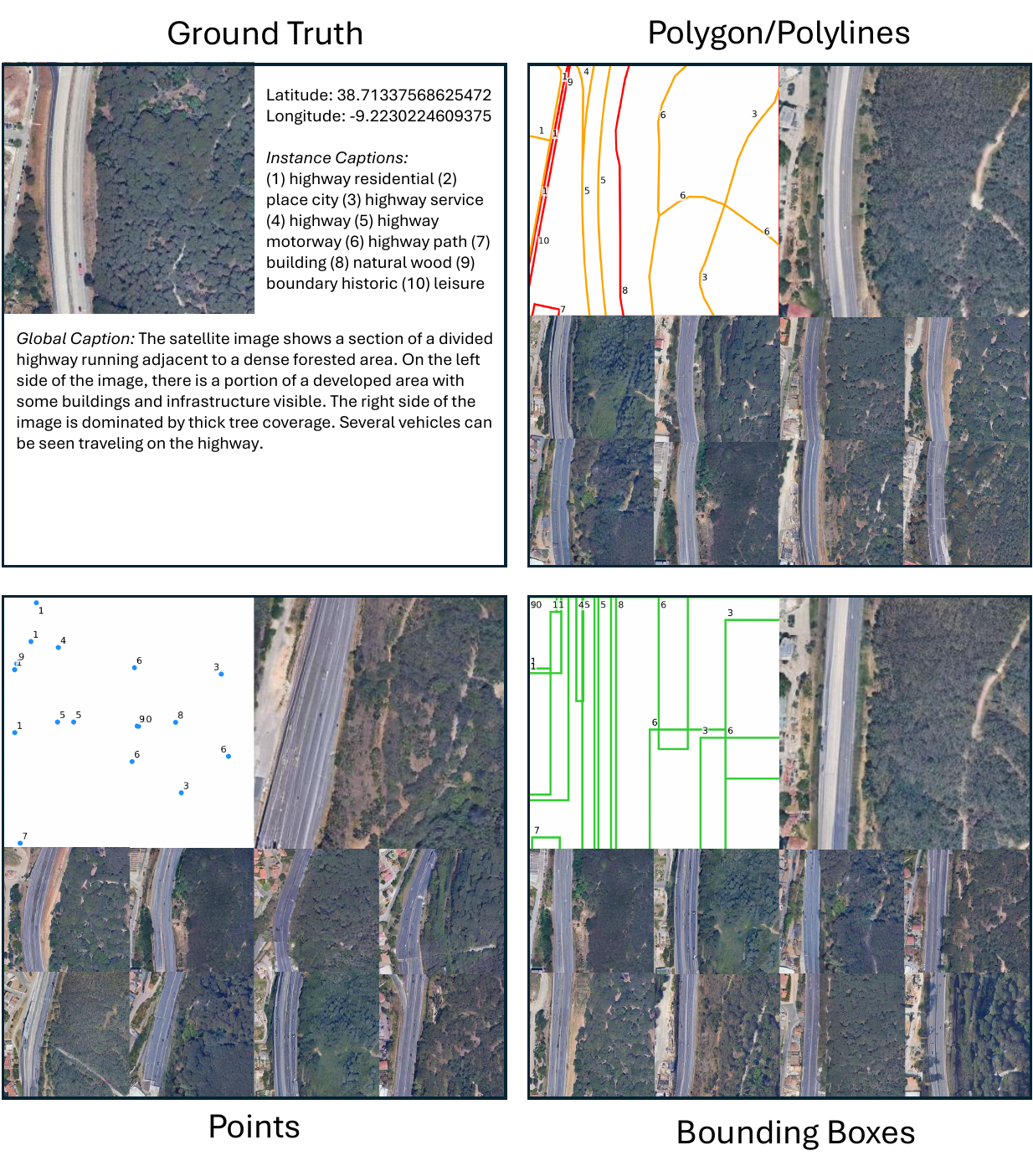}
    \captionof{figure}{\TerraDiTOmega{}-XL Uncurated Samples. Highway and forest in Lisbon, Portugal.}
    \label{fig:samples_2}
\end{center}

\begin{figure}[tb]
    \centering
    \includegraphics[width=0.6\linewidth]{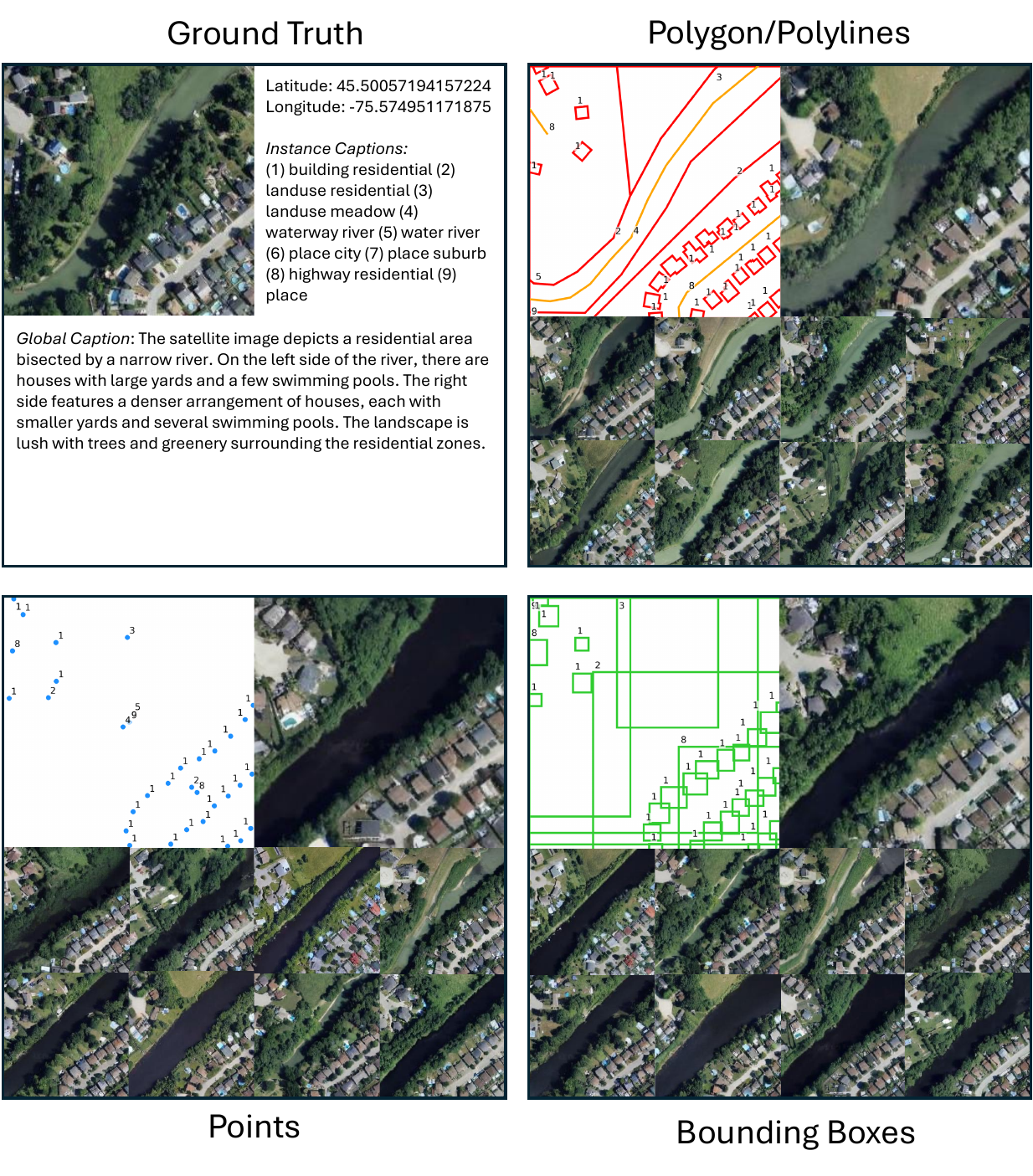}
    \caption{\TerraDiTOmega{}-XL Uncurated Samples. Residential and river in Quebec, Canada.}
    \label{fig:samples_3}

    \par\medskip

    \includegraphics[width=0.6\linewidth]{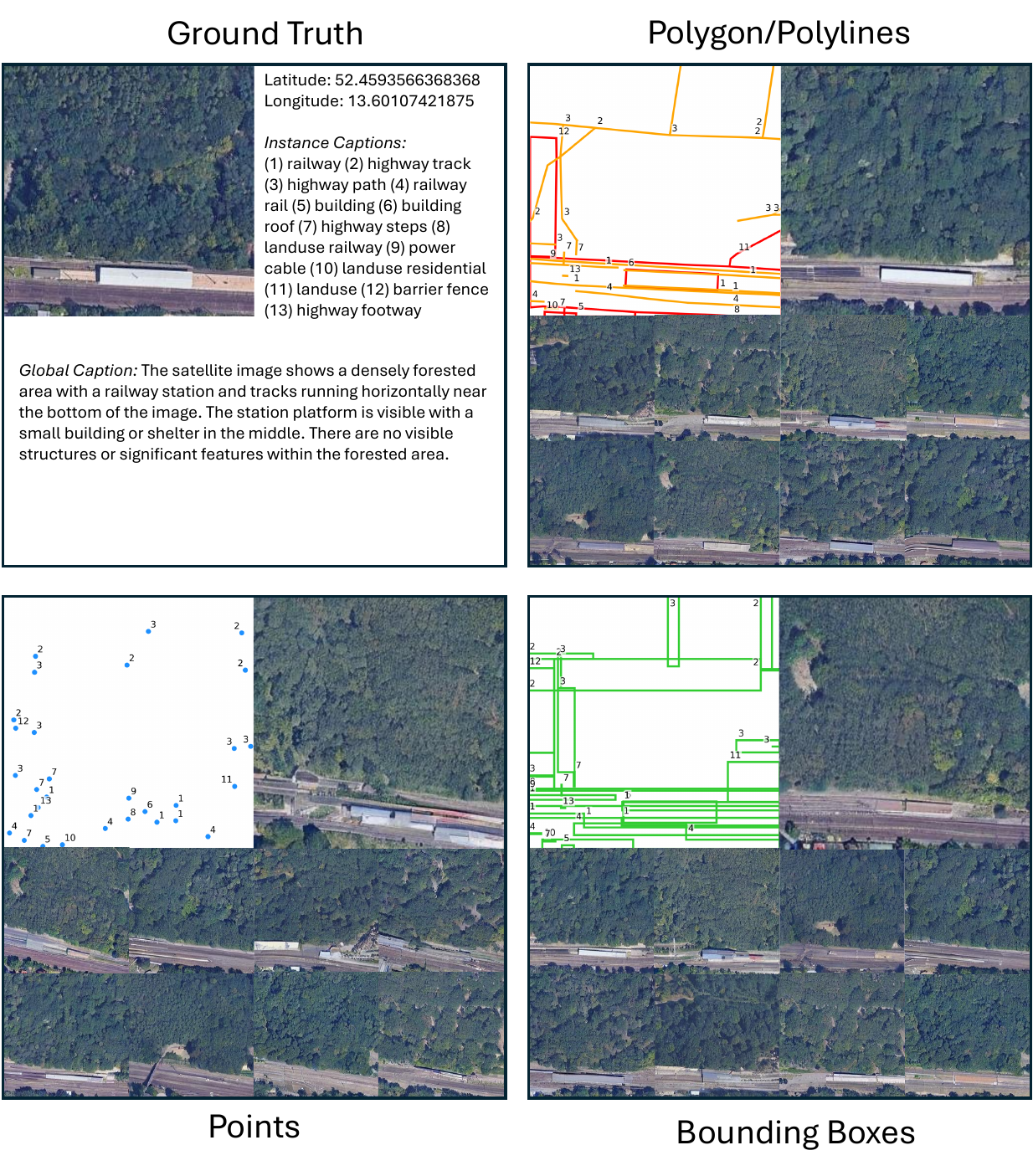}
    \caption{\TerraDiTOmega{}-XL Uncurated Samples. Railway and forest in Berlin, Germany.}
    \label{fig:samples_4}
\end{figure}

\end{document}